\documentclass{article} 
\usepackage{iclr2027_conference,times}

\usepackage{amsmath,amsfonts,bm}

\def\eqref#1{equation~\ref{#1}}

\def\1{\bm{1}}

\DeclareMathAlphabet{\mathsfit}{\encodingdefault}{\sfdefault}{m}{sl}
\SetMathAlphabet{\mathsfit}{bold}{\encodingdefault}{\sfdefault}{bx}{n}

\newcommand{\E}{\mathbb{E}}

\newcommand{\R}{\mathbb{R}}

\usepackage{hyperref}
\usepackage{url}

\usepackage{amsmath}
\usepackage{amssymb}
\usepackage{mathtools}
\usepackage{amsthm}
\usepackage{adjustbox}
\usepackage{enumitem}

\usepackage{wrapfig}
\usepackage{booktabs}
\usepackage{thm-restate}
\usepackage{multirow}

\theoremstyle{plain}

\theoremstyle{definition}

\theoremstyle{remark}
\newtheorem{example}{Example}[section]

\usepackage{tikz}
\usepackage{pgfplots}
\usetikzlibrary{matrix,fit,calc,arrows.meta,decorations.pathreplacing, pgfplots.groupplots}
\pgfmathdeclarefunction{qsquartet}{2}{%
  \pgfmathparse{(20.460661726679252*(0.65283764775909348*#2)^(-0.62178775679245868)
    +2.7950430030333582*(0.8746198315994006*#1*#2/1000)^(-0.55016534555575614))^(0.32212913692610312)
    +1.6501142610078894+0}%
}
\pgfmathdeclarefunction{qsours}{2}{%
  \pgfmathparse{(20.460661726679252*(0.77007787288087171*#2)^(-0.62178775679245868)
    +2.7950430030333582*(0.95095179660011742*#1*#2/1000)^(-0.55016534555575614))^(0.32212913692610312)
    +1.6501142610078894+0.041587153424023846}%
}
\newcommand{\QuantizedScalingPlot}[1][]{%
\begin{tikzpicture}[baseline=(scalingaxis.south)]
\begin{axis}[
  name=scalingaxis,
  width=0.86\linewidth,
  height=0.60\linewidth,
  scale only axis,
  xmode=log, log basis x=2,
  ymode=log,
  xmin=21, xmax=115,
  ymin=2.78, ymax=3.00,
  xtick={25,50,100},
  xticklabels={25,50,100},
  ytick={2.8,2.85,2.9,2.95,3.0},
  log ticks with fixed point,
  xlabel={Tokens per parameter},
  ylabel={Validation loss},
  axis lines*=left,
  axis line style={black,line width=0.65pt},
  tick align=outside,
  tick style={black,line width=0.5pt},
  minor tick num=0,
  grid=major,
  grid style={black!10,line width=0.3pt},
  label style={font=\small},
  tick label style={font=\footnotesize},
  legend columns=2,
  legend style={
    /tikz/every even column/.append style={column sep=12pt},
    at={(0.98,0.98)},
    anchor=north east,
    draw=black!70,
    fill=white,
    line width=0.4pt,
    inner sep=2pt,
    font=\tiny,
    cells={anchor=west},
    row sep=0pt
  },
  legend image post style={scale=0.7},
  samples=160,
  clip=true,
  clip mode=individual,
  #1
]

\addplot[
  red,dashed,line width=0.6pt,
  no marks,
  domain=24.607066857334:98.429557581217
] {qsquartet(x,203.188480)};
\addlegendentry{Quartet}

\addplot[
  red,solid,line width=0.6pt,
  no marks,
  domain=24.607066857334:98.429557581217
] {qsours(x,203.188480)};
\addlegendentry{Ours}

\addplot[
  only marks,
  mark=diamond*,
  mark size=2.8pt,
  mark options={fill=red,draw=red}
] coordinates {
  (24.607066857334,2.96302)
  (49.214133714667,2.87813)
  (98.429557581217,2.81429)
};
\addlegendentry{\makebox[0pt][l]{Held-out (200M)}}

\end{axis}
\end{tikzpicture}%
}

\pgfmathdeclarefunction{qcalquartet}{2}{%
  \pgfmathparse{(20.460661726679252*(0.65283764775909348*#2)^(-0.62178775679245868)
    +2.7950430030333582*(0.8746198315994006*#1*#2/1000)^(-0.55016534555575614))^(0.32212913692610312)
    +1.6501142610078894+0}%
}
\pgfmathdeclarefunction{qcalours}{2}{%
  \pgfmathparse{(20.460661726679252*(0.77007787288087171*#2)^(-0.62178775679245868)
    +2.7950430030333582*(0.95095179660011742*#1*#2/1000)^(-0.55016534555575614))^(0.32212913692610312)
    +1.6501142610078894+0.041587153424023846}%
}
\newcommand{\QuantizedScalingCalibrationPlot}[1][]{%
\begin{tikzpicture}[baseline=(calibrationaxis.south)]
\begin{axis}[
  name=calibrationaxis,
  width=0.86\linewidth,
  height=0.60\linewidth,
  scale only axis,
  xmode=log, log basis x=2,
  ymode=log,
  xmin=21, xmax=1500,
  ymin=2.82, ymax=3.60,
  xtick={25,50,100,200,400,800},
  xticklabels={25,50,100,200,400,800},
  ytick={2.9,3.1,3.3,3.5},
  log ticks with fixed point,
  xlabel={Tokens per parameter},
  ylabel={Validation loss},
  axis lines*=left,
  axis line style={black,line width=0.65pt},
  tick align=outside,
  tick style={black,line width=0.5pt},
  minor tick num=0,
  grid=major,
  grid style={black!10,line width=0.3pt},
  label style={font=\small},
  tick label style={font=\footnotesize},
  legend columns=2,
  legend style={
    /tikz/every even column/.append style={column sep=12pt},
    at={(0.98,0.98)},
    anchor=north east,
    draw=black!70,
    fill=white,
    line width=0.4pt,
    inner sep=2pt,
    font=\tiny,
    cells={anchor=west},
    row sep=0pt
  },
  legend image post style={scale=0.7},
  legend image post style={mark size=2.8pt},
  samples=160,
  clip=true,
  clip mode=individual,
  #1
]

\addlegendimage{blue,dashed,line width=0.5pt}
\addlegendentry{Quartet}
\addlegendimage{blue,solid,line width=0.5pt}
\addlegendentry{Ours}

\addplot[
  blue,dashed,line width=0.5pt,
  no marks,forget plot,
  variable=\t,domain=4.620823049132:9.620823049132
] ({2^\t},{qcalquartet(2^\t,30.482560)});
\addplot[
  blue,solid,line width=0.5pt,
  no marks,forget plot,
  variable=\t,domain=4.620823049132:9.620823049132
] ({2^\t},{qcalours(2^\t,30.482560)});
\addplot[
  blue,dashed,line width=0.5pt,
  no marks,forget plot,
  variable=\t,domain=4.656590687168:9.656694694832
] ({2^\t},{qcalquartet(2^\t,49.556736)});
\addplot[
  blue,solid,line width=0.5pt,
  no marks,forget plot,
  variable=\t,domain=4.656590687168:9.656694694832
] ({2^\t},{qcalours(2^\t,49.556736)});
\addplot[
  blue,dashed,line width=0.5pt,
  no marks,forget plot,
  variable=\t,domain=4.604214299169:9.604323034277
] ({2^\t},{qcalquartet(2^\t,102.777856)});
\addplot[
  blue,solid,line width=0.5pt,
  no marks,forget plot,
  variable=\t,domain=4.604214299169:9.604323034277
] ({2^\t},{qcalours(2^\t,102.777856)});

\addplot[
  only marks,
  mark=diamond*,
  mark size=1.4pt,
  mark options={fill=blue,draw=blue}
] coordinates {
  (24.604035356610,3.50033)
  (49.208070713221,3.38181)
  (98.416141426442,3.29879)
  (196.832282852884,3.24354)
  (393.664565705767,3.20481)
  (787.329131411535,3.16667)
};
\addlegendentry{\makebox[0pt][l]{Calibration (30--100M)}}
\addplot[
  only marks,
  mark=diamond*,
  mark size=1.4pt,
  mark options={fill=blue,draw=blue},
  forget plot
] coordinates {
  (25.221648818841,3.32075)
  (50.443297637681,3.22408)
  (100.891885050702,3.15832)
  (201.783770101405,3.10953)
  (403.572829978149,3.07157)
  (807.150949731637,3.03485)
};
\addplot[
  only marks,
  mark=diamond*,
  mark size=1.4pt,
  mark options={fill=blue,draw=blue},
  forget plot
] coordinates {
  (24.322410305971,3.13147)
  (48.647371200271,3.04042)
  (97.294742400542,2.97248)
  (194.592035389413,2.92403)
  (389.186621367155,2.88647)
  (778.375793322639,2.85331)
};

\node[anchor=west,font=\tiny,text=black!80] at (axis cs:905.428501,3.16667) {30M};
\node[anchor=west,font=\tiny,text=black!80] at (axis cs:928.223592,3.03485) {50M};
\node[anchor=west,font=\tiny,text=black!80] at (axis cs:895.132162,2.85331) {100M};

\end{axis}
\end{tikzpicture}%
}

\newcommand{\MixedParityPlot}[1][]{%
\begin{tikzpicture}
\begin{axis}[
  name=ax0, 
  width=0.26\linewidth, height=0.26\linewidth, scale only axis,
  xmin=3.0260, xmax=3.1403, ymin=3.0260, ymax=3.1403,
  xtick={3.04,3.06,3.08,3.1,3.12,3.14}, ytick={3.04,3.06,3.08,3.1,3.12,3.14},
  xticklabel style={/pgf/number format/.cd, fixed, fixed zerofill, precision=2},
  yticklabel style={/pgf/number format/.cd, fixed, fixed zerofill, precision=2},
  scaled ticks=false,
  title={Ours}, title style={font=\small, text height=1.6ex, text depth=0.3ex, yshift=-4pt},
  xlabel={Observed loss}, ylabel={Predicted loss},
  axis line style={black, line width=0.65pt},
  tick align=outside, tick style={black, line width=0.5pt},
  xtick pos=bottom, ytick pos=left,
  grid=major, grid style={black!10, line width=0.3pt},
  label style={font=\small}, tick label style={font=\scriptsize},
  clip=false,
  legend columns=1,
  legend style={at={(2.2331,0.5)}, anchor=west, draw=black!70, fill=white,
    line width=0.4pt, font=\footnotesize, inner sep=2pt,
    row sep=1pt, cells={anchor=west}},
  legend image post style={mark size=2.4pt},
  #1
]
\addplot[black, dashed, line width=0.7pt, forget plot] coordinates {(3.0260,3.0260) (3.1403,3.1403)};
\addplot[only marks, mark size=1.4pt, mark options={mark=*,        fill=blue,   draw=black, line width=0.5pt}] coordinates {(3.11771,3.11787) (3.08733,3.08993) (3.06543,3.06838) (3.04446,3.04733) (3.11319,3.11197) (3.08286,3.08400) (3.06231,3.06242) (3.04096,3.04135) (3.11050,3.10641) (3.08062,3.07840) (3.05878,3.05680) (3.03816,3.03570)};
\addlegendentry{Forward, first/last}
\addplot[only marks, mark size=1.4pt, mark options={mark=square*,  fill=cyan,   draw=black, line width=0.5pt}] coordinates {(3.11876,3.11826) (3.08938,3.09032) (3.06889,3.06878) (3.04818,3.04773) (3.11255,3.11266) (3.08244,3.08469) (3.06155,3.06311) (3.04061,3.04205) (3.10713,3.10704) (3.07666,3.07904) (3.05552,3.05744) (3.03437,3.03635)};
\addlegendentry{Forward, middle}
\addplot[only marks, mark size=1.4pt, mark options={mark=triangle*, fill=red,   draw=black, line width=0.5pt}] coordinates {(3.12654,3.12973) (3.09752,3.10181) (3.07650,3.08028) (3.05553,3.05925) (3.12113,3.12270) (3.09129,3.09477) (3.07066,3.07323) (3.04958,3.05219) (3.12017,3.11773) (3.08980,3.08979) (3.06858,3.06824) (3.04731,3.04720)};
\addlegendentry{Backward, first/last}
\addplot[only marks, mark size=1.4pt, mark options={mark=triangle*, fill=orange, draw=black, rotate=180, line width=0.5pt}] coordinates {(3.12854,3.13415) (3.09931,3.10623) (3.07820,3.08470) (3.05837,3.06368) (3.12943,3.13052) (3.09996,3.10259) (3.07927,3.08106) (3.05845,3.06003) (3.12460,3.12562) (3.09586,3.09769) (3.07566,3.07615) (3.05542,3.05512)};
\addlegendentry{Backward, middle}
\addplot[only marks, mark size=1.2pt, mark options={mark=*,        fill=blue,   draw=black, line width=0.5pt}, forget plot] coordinates {(3.11771,3.11787) (3.08733,3.08993) (3.06543,3.06838) (3.04446,3.04733) (3.11319,3.11197) (3.08286,3.08400) (3.06231,3.06242) (3.04096,3.04135) (3.11050,3.10641) (3.08062,3.07840) (3.05878,3.05680) (3.03816,3.03570)};
\node[anchor=north west, draw=black!70, fill=white, font=\scriptsize, align=left, inner sep=2pt] at (rel axis cs:0.02,0.985) {$R^2 = 0.991$\\ $\mathrm{MSE} = 7.1\times10^{-6}$};
\end{axis}
\begin{axis}[
  name=ax1, at={(ax0.south east)}, anchor=south west, xshift=0.045\linewidth,
  width=0.26\linewidth, height=0.26\linewidth, scale only axis,
  xmin=3.0260, xmax=3.1403, ymin=3.0260, ymax=3.1403,
  xtick={3.04,3.06,3.08,3.1,3.12,3.14}, ytick={3.04,3.06,3.08,3.1,3.12,3.14},
  xticklabel style={/pgf/number format/.cd, fixed, fixed zerofill, precision=2},
  yticklabel style={/pgf/number format/.cd, fixed, fixed zerofill, precision=2},
  scaled ticks=false,
  title={Quartet}, title style={font=\small, text height=1.6ex, text depth=0.3ex, yshift=-4pt},
  xlabel={Observed loss}, yticklabels={},
  axis line style={black, line width=0.65pt},
  tick align=outside, tick style={black, line width=0.5pt},
  xtick pos=bottom, ytick pos=left,
  grid=major, grid style={black!10, line width=0.3pt},
  label style={font=\small}, tick label style={font=\scriptsize},
  clip=false,
  #1
]
\addplot[black, dashed, line width=0.7pt, forget plot] coordinates {(3.0260,3.0260) (3.1403,3.1403)};
\addplot[only marks, mark size=1.4pt, mark options={mark=*,        fill=blue,   draw=black, line width=0.5pt}] coordinates {(3.11771,3.11385) (3.08733,3.08525) (3.06543,3.06321) (3.04446,3.04167) (3.11319,3.10889) (3.08286,3.08022) (3.06231,3.05810) (3.04096,3.03651) (3.11050,3.10472) (3.08062,3.07598) (3.05878,3.05381) (3.03816,3.03216)};
\addplot[only marks, mark size=1.4pt, mark options={mark=square*,  fill=cyan,   draw=black, line width=0.5pt}] coordinates {(3.11876,3.11421) (3.08938,3.08563) (3.06889,3.06358) (3.04818,3.04206) (3.11255,3.10947) (3.08244,3.08081) (3.06155,3.05871) (3.04061,3.03712) (3.10713,3.10518) (3.07666,3.07646) (3.05552,3.05429) (3.03437,3.03265)};
\addplot[only marks, mark size=1.4pt, mark options={mark=triangle*, fill=red,   draw=black, line width=0.5pt}] coordinates {(3.12654,3.12412) (3.09752,3.09504) (3.07650,3.07262) (3.05553,3.05072) (3.12113,3.11665) (3.09129,3.08828) (3.07066,3.06640) (3.04958,3.04504) (3.12017,3.11218) (3.08980,3.08424) (3.06858,3.06269) (3.04731,3.04165)};
\addplot[only marks, mark size=1.4pt, mark options={mark=triangle*, fill=orange, draw=black, rotate=180, line width=0.5pt}] coordinates {(3.12854,3.12918) (3.09931,3.09963) (3.07820,3.07683) (3.05837,3.05457) (3.12943,3.12491) (3.09996,3.09576) (3.07927,3.07328) (3.05845,3.05132) (3.12460,3.11951) (3.09586,3.09087) (3.07566,3.06878) (3.05542,3.04722)};
\addplot[only marks, mark size=1.2pt, mark options={mark=*,        fill=blue,   draw=black, line width=0.5pt}, forget plot] coordinates {(3.11771,3.11385) (3.08733,3.08525) (3.06543,3.06321) (3.04446,3.04167) (3.11319,3.10889) (3.08286,3.08022) (3.06231,3.05810) (3.04096,3.03651) (3.11050,3.10472) (3.08062,3.07598) (3.05878,3.05381) (3.03816,3.03216)};
\node[anchor=north west, draw=black!70, fill=white, font=\scriptsize, align=left, inner sep=2pt] at (rel axis cs:0.02,0.985) {$R^2 = 0.974$\\ $\mathrm{MSE} = 2.0\times10^{-5}$};
\end{axis}
\end{tikzpicture}%
}

\usepackage{capt-of}
\theoremstyle{remark}

\usepackage{xcolor}
\usepackage{pifont}

\newcommand{\cmark}{\textcolor{green!60!black}{\large \ding{51}}}
\newcommand{\xmark}{\textcolor{red!70!black}{\large \ding{55}}}

\newcommand{\Lcal}{\mathcal{L}}

\usepackage{xcolor}
\usepackage{graphicx}
\usepackage{wrapfig}
\usepackage{booktabs}
\usepackage{adjustbox}
\title{Mixed Precision Scaling Laws for Quantized Pretraining}

\author{%
Jinuk Kim$^\text{1}$\textsuperscript{\dag}\thanks{Work done during internship at Amazon. ~ \textsuperscript{\dag}Correspondence to: Jinuk Kim \texttt{<jusjinuk@snu.ac.kr>} and Youngsuk Park \texttt{<pyoungsu@amazon.com>}}~,
\enspace Rajarshi Saha$^\text{2}$,
\enspace Muhammed Fatih Bal{\i}n$^\text{2}$,
\enspace Youngsuk Park$^\text{2}$\textsuperscript{\dag}
\\[2pt]
\textsuperscript{1}Seoul National University
\textsuperscript{2}Amazon Annapurna Labs
}

\iclrfinalcopy 
\begin{document}

\maketitle
\lhead{Preprint}

\begin{abstract}
Since scaling laws were introduced for modern LLMs \citep{kaplan2020scaling}, several works have extended them to account for quantization, tracking the emerging low-precision data types supported on AI accelerators. We develop a more accurate scaling law for quantized pretraining with mixed precision across the forward and backward passes, grounded in optimization theory and validated empirically. Our formulation separates the effects of quantization on model capacity and on optimization efficiency, and explicitly models the persistent optimization error from forward and backward quantization identified through convergence analysis. To select the regimes for fitting and evaluating our law, we first decompose existing 4-bit training recipes into their forward and backward components, isolating those that most affect loss. Our Qwen3 experiments under selected 4-bit regimes, together with published Llama-style experiments, show that our law predicts loss with up to $11\times$ lower MSE than existing laws as model size and token count scale. Finally, we extend our law to layer-wise mixed precision and identify the optimal quantized-pretraining recipe, i.e., model size, training tokens, and per-layer forward/backward precision, under a hardware-aware compute budget. Unlike prior precision scaling laws, our law predicts that the optimal precision increases with the compute budget.
\end{abstract}

\section{Introduction}

\begin{figure*}[t]
\centering
\includegraphics[width=0.8\textwidth]{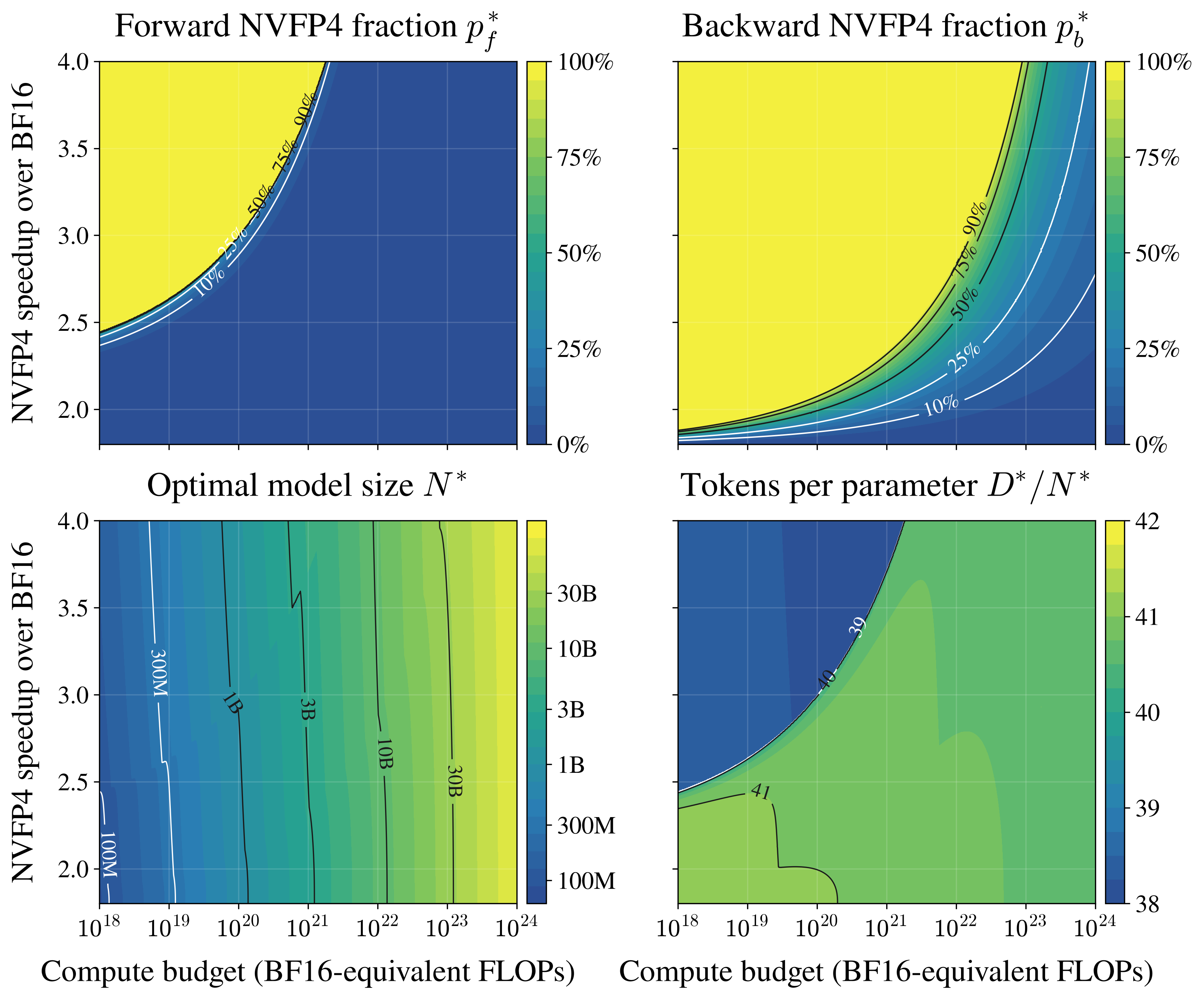}
\caption{Compute-optimal precision allocation predicted by our scaling law. With each Transformer block using MXFP8 or NVFP4 in each pass, the panels show the optimal fraction of NVFP4 blocks in the forward pass $p_f^*$ (top, left) and backward pass $p_b^*$ (top, right), the optimal model size $N^*$ (bottom, left), and tokens per parameter $D^*/N^*$ (bottom, right), given the compute budget ($x$-axis) and the NVFP4 speedup over BF16 ($y$-axis), with the MXFP8 speedup fixed at $1.8\times$ (Section~\ref{sec:compute-optimal}).}
\vspace{-1em}
\label{fig:compute-optimal}
\end{figure*}

Scaling laws have played a central role in guiding large language model (LLM) pretraining by characterizing how training loss varies with model size, data, and compute \citep{kaplan2020scaling,hoffmann2022training,videau2026skaling}. By extrapolating from smaller scale experiments, they enable researchers to choose effective training configurations for expensive runs in advance. As pretraining costs continue to grow, accurate scaling laws are becoming increasingly valuable for allocating limited resources and maximizing model quality under a given compute budget.

Quantized training offers a promising path towards more efficient LLM pretraining by performing computations in the forward and backward passes at reduced numerical precision \citep{micikevicius2022fp8,fishman2025scalingfp8}. Recent recipes push precision down to 4 bits using block-scaled formats such as NVFP4, while combining different techniques in the forward and backward passes to preserve training quality \citep{tseng2025mxfp4,nvidia2025nvfp4}. Existing methods differ both in how quantization scales are shared and selected during the forward pass and in how gradients are estimated during the backward pass. For example, \citet{nvidia2025nvfp4} and \citet{panferov2026quartetii} adopt different choices along both dimensions. Together, these techniques have substantially narrowed the gap between FP4 and BF16 training, making low-precision training an increasingly practical option for LLM pretraining \citep{chmiel2025fp4alltheway,nvidia2025nvfp4}.

We first revisit existing NVFP4 training recipes by decomposing them into forward and backward components and isolating the contribution of each to pretraining loss, since prior works typically proposed and evaluated these techniques only as complete configurations (Section~\ref{sec:recipes}). We find that forward quantization choices affect loss more than backward techniques, but even the best combination leaves a noticeable loss gap relative to BF16 training. Mixed-precision quantized training offers a promising way to narrow this gap, either by assigning different precisions to the forward and backward passes \citep{castro2025quartet} or by varying precision across layers \citep{nvidia2025nvfp4}. 
Choosing among such configurations before training requires accurately predicting their loss at the target model size and token budget, which motivates scaling laws for mixed precision both across the forward and backward passes and across layers.

For mixed precision across the forward and backward passes, existing scaling laws model forward quantization as a reduction in the effective parameter count \citep{kumar2024scalingprecision,frantar2025compression,panferov2026unified}, and the Quartet law additionally models backward quantization as a reduction in the effective training token count, whose penalty can be offset by training on more data \citep{castro2025quartet} (Table~\ref{tab:scaling-law-formulations}). However, quantization also introduces optimization error that can persist with continued training. In the forward pass, evaluating gradients at quantized weights while updating full-precision weights can induce oscillations that prevent convergence \citep{bengio2013estimating,bai2018proxquant,defossez2021differentiable,nagel2022overcoming}, and in the backward pass, quantization can bias gradient estimates \citep{ajalloeian2020convergence}. Accurate loss prediction therefore requires modeling this persistent error in both passes.

In this work, we develop a scaling law for forward--backward mixed-precision pretraining that accounts for this persistent optimization error (Section~\ref{sec:revised}). Guided by a theoretical analysis of the optimization gap, the law combines effective parameter and data counts with separate additive terms for persistent optimization error from forward and backward quantization. The resulting law (Table~\ref{tab:scaling-law-formulations}) takes the form
\begin{equation*}
L = \left( A\left(\rho_N(\psi_{f})N\right)^{-\alpha} + B\left(\rho_D(\psi_{b})D\right)^{-\beta} \right)^{\gamma}
+ \epsilon_{f}(\psi_{f}) + \epsilon_{b}(\psi_{b}) + E,
\end{equation*}
where $\rho_N$ and $\rho_D$ rescale the parameter count $N$ and the training token count $D$, respectively, and $\epsilon_{f}$ and $\epsilon_{b}$ model persistent optimization error associated with the forward and backward quantization schemes $\psi_{f}$ and $\psi_{b}$, respectively. We evaluate the law on our Qwen3 \citep{qwen3} pretraining experiments, covering models with up to 0.9B parameters trained on up to 35.8B tokens with BF16, MXFP8, and different NVFP4 schemes in both passes, and on the published Quartet results for Llama-style models trained with up to 800 tokens per parameter \citep{castro2025quartet}. Across both settings, the law predicts held-out loss more accurately than existing laws when extrapolating to larger model sizes and token budgets.


For mixed precision across layers, we extend the law to predict the loss when different quantization schemes are assigned to different Transformer blocks (Section~\ref{sec:mixed}). The extension interpolates between the predicted losses of the two schemes being mixed, with an exponent that captures the strategy used to select blocks for each scheme (Table~\ref{tab:scaling-law-formulations}). Accounting for the hardware speedup of each format, we then use the law to identify the compute-optimal model size, number of training tokens, and precision of the forward and backward passes under a given compute budget (Figure~\ref{fig:compute-optimal}). 

We summarize our main contributions as follows:
\begin{enumerate}[label=(\alph*), leftmargin=*, itemsep=0pt]

    \item We revisit existing NVFP4 recipes by decomposing them into forward and backward components, identifying which components matter for loss and measuring the remaining gap to BF16, which guides the design of our scaling law experiments (Section~\ref{sec:recipes}).

    \item We develop a scaling law that models persistent optimization error from forward and backward quantization under mixed precision across the two passes, backed by theoretical analysis (Table~\ref{tab:scaling-law-formulations}). It predicts held-out loss more accurately than existing laws on our Qwen3 experiments and the published Quartet results (Section~\ref{sec:revised}).

    \item We extend the law to layer-wise mixed precision, and identify compute-optimal model sizes and precision assignments under a hardware-aware compute budget (Section~\ref{sec:mixed}).
\end{enumerate}

\section{Revisiting quantized pretraining recipes}
\label{sec:recipes}

\begin{wraptable}{r}{0.53\linewidth}
\vspace{-2em}
\centering
\caption{Validation loss of FP4 pretraining recipes decomposed into forward and backward components, for a 430M parameter Qwen3 model trained at 12.8B tokens. Loss increase is relative to the BF16 baseline. The last column indicates whether each combination is included in our scaling law grid.}
\label{tab:recipe-ablation}
\vspace{0.5em}
\begin{adjustbox}{max width=\linewidth}
\begin{tabular}{llccc}
\toprule
Forward & Backward &
Val. loss $\downarrow$ &
Increase $\downarrow$ &
In grid \\
\cmidrule(lr){1-1}\cmidrule(lr){2-2}\cmidrule(lr){3-4}\cmidrule(lr){5-5}
BF16 & BF16 &
3.245 & \multicolumn{1}{c}{--} & \cmark \\
\cmidrule(lr){2-2}\cmidrule(lr){3-4}\cmidrule(lr){5-5}
 & MS-EDEN &
3.257 & +0.36\% & \cmark \\
\cmidrule(lr){2-2}\cmidrule(lr){3-4}\cmidrule(lr){5-5}
 & RHT + SR &
3.255 & +0.32\% & \cmark \\
\cmidrule(lr){2-2}\cmidrule(lr){3-4}\cmidrule(lr){5-5}
& SR &
3.268 & +0.72\% & \cmark \\
\cmidrule(lr){1-1}\cmidrule(lr){2-2}\cmidrule(lr){3-4}\cmidrule(lr){5-5}
4/6 + 1D scaling & BF16 &
3.275 & +0.94\% & \cmark \\
\cmidrule(lr){2-2}\cmidrule(lr){3-4}\cmidrule(lr){5-5}
& MS-EDEN &
3.280 & +1.07\% & \cmark \\
\cmidrule(lr){2-2}\cmidrule(lr){3-4}\cmidrule(lr){5-5}
& RHT + SR &
3.282 & +1.14\% & \cmark \\
\cmidrule(lr){2-2}\cmidrule(lr){3-4}\cmidrule(lr){5-5}
& SR &
3.291 & +1.42\% & \cmark \\
\cmidrule(lr){1-1}\cmidrule(lr){2-2}\cmidrule(lr){3-4}\cmidrule(lr){5-5}
2D scaling & BF16 &
3.284 & +1.22\% & \xmark \\
\cmidrule(lr){2-2}\cmidrule(lr){3-4}\cmidrule(lr){5-5}
 & RHT + SR &
3.291 & +1.41\% & \xmark \\
\cmidrule(lr){2-2}\cmidrule(lr){3-4}\cmidrule(lr){5-5}
& SR &
3.300 & +1.69\% & \xmark \\
\bottomrule
\end{tabular}
\end{adjustbox}
\vspace{-1em}
\end{wraptable}

Quantized pretraining reduces training cost by quantizing computation in both the forward and backward passes, with recent recipes pushing precision down to FP4 using block-scaled formats such as NVFP4. Since these recipes are typically evaluated as complete configurations, it remains unclear which of their individual forward and backward components matter for pretraining loss. 
In this section, we decompose two state-of-the-art FP4 recipes, the NVFP4 recipe \citep{nvidia2025nvfp4} and Quartet II \citep{panferov2026quartetii}, into their forward and backward components to isolate which matter most for pretraining loss, and select the most effective regimes to adopt in our scaling law experiments.
In the forward pass, the NVFP4 recipe uses \textit{2D scaling}, sharing scales across $16\times16$ weight blocks so that quantized weights remain consistent when transposed in the backward pass, whereas Quartet II uses \textit{1D scaling} with $1\times16$ blocks and \textit{4/6 scaling}, which selects the lower-error of two candidate scales for each block \citep{cook2025fouroversix}. In the backward pass, the NVFP4 recipe applies \textit{stochastic rounding (SR)} to the output activation gradients and \textit{randomized Hadamard transforms (RHT)} to the operands of the weight gradient computation, similar to the MXFP4 recipe \citep{tseng2025mxfp4}, whereas Quartet II uses \textit{MS-EDEN}, which combines randomized rotations with EDEN-based rescaling to obtain unbiased matrix multiplication.

\paragraph{Results}

Table~\ref{tab:recipe-ablation} reports the validation loss of a Qwen3 model \citep{qwen3} with 430M non-embedding parameters trained on 12.8B tokens for each combination of forward and backward components. The results show that forward pass quantization is generally more sensitive than backward pass quantization. From the BF16 baseline, quantizing only the backward pass to NVFP4 increases the loss by 0.32--0.72\%, whereas quantizing only the forward pass increases it by 0.94--1.22\%. Similarly, the advantage of Quartet II (4/6 + 1D scaling, MS-EDEN) over the NVFP4 recipe (2D scaling, RHT + SR) comes largely from the forward pass. Starting from Quartet II, replacing MS-EDEN with RHT + SR increases the loss by only 0.07 percentage points, whereas subsequently replacing 4/6 + 1D scaling with 2D scaling while keeping RHT + SR in the backward pass increases it by a further 0.27 percentage points. Based on these results, we adopt 4/6 + 1D scaling for NVFP4 forward quantization rather than 2D scaling, which performs worse overall, and retain all three backward techniques in our scaling law experiments. These selected regimes, together with BF16 and MXFP8 in both passes, form our experimental grid (Table~\ref{tab:dataset-overview}).

Notably, even with the best FP4 recipe (Quartet II), a noticeable loss gap remains relative to the BF16 baseline, suggesting that uniformly quantizing to 4-bit may not be preferable. Mixed precision offers a way to narrow this gap by assigning different precisions to the forward and backward passes \citep{castro2025quartet} or by varying precision across layers \citep{nvidia2025nvfp4,li2026nvfp4rl}. Accurately predicting the quantized pretraining loss of these configurations can help practitioners choose ones that meet their quality and compute targets, motivating quantized pretraining scaling laws for mixed precision across the forward and backward passes and across layers.

To this end, we develop a scaling law for mixed precision across the forward and backward passes in Section~\ref{sec:revised}, guided by theoretical analysis, that predicts quantized pretraining loss more accurately than existing laws. In Section~\ref{sec:mixed}, we extend this law to layer-wise mixed precision and use it to identify compute-optimal configurations. We evaluate these laws on the selected regimes, and additionally validate the law in Section~\ref{sec:revised} on the published Quartet grid \citep{castro2025quartet}.

\section{Forward--backward mixed-precision scaling law with optimization error}

\label{sec:revised}

\begin{table*}[t]

\caption{Comparison of scaling laws for quantization and compression. The last three columns indicate whether each law models different schemes in the forward and backward passes, different schemes across layers, and persistent optimization error. $\rho_N$ and $\rho_D$ rescale $N$ and $D$ according to the forward and backward schemes $\psi_f$ and $\psi_b$, and $\epsilon_f$ and $\epsilon_b$ model persistent optimization error independent of $N$ and $D$. For layer-wise mixed precision, $p \in [0,1]$ is the fraction of layers assigned to the second of two schemes with losses $L_1$ and $L_2$, and $\gamma_S$ depends on the layer selection strategy $S$. PTQ, QAT, and QPT denote post-training quantization, quantization-aware training, and quantized pretraining, respectively. Other notation is described in Appendix~\ref{sec:app-related}.}
\label{tab:scaling-law-formulations}
\vspace{0.5em}
\centering
\begin{adjustbox}{max width=\textwidth}
\begin{tabular}{llccc}
\toprule
&
& {\bf Fwd--bwd} & {\bf Layer} & {\bf Optim.} \\
{\bf Scaling law} & {\bf Loss formulation}
& {\bf mixed} & {\bf mixed} & {\bf error} $\epsilon(\cdot)$ \\
\cmidrule(lr){1-1}\cmidrule(lr){2-2}\cmidrule(lr){3-5}
Precision (PTQ) \citep{kumar2024scalingprecision}
& $L=AN^{-\alpha}+BD^{-\beta}+E
+C_T D^{\gamma_D}N^{-\gamma_N}
e^{-P_{\mathrm{post}}/\gamma_{\mathrm{post}}}$
& \xmark & \xmark & \xmark \\
\cmidrule(lr){1-1}\cmidrule(lr){2-2}\cmidrule(lr){3-5}
Mixed (PTQ) \citep{cao2024mixedquant}
& $L=AN^{-\alpha}+BD^{-\beta}+E
+C e^{\kappa p}N^{-\gamma_N}(G+d)^{\gamma_G}$
& \xmark & \cmark & \xmark \\
\cmidrule(lr){1-1}\cmidrule(lr){2-2}\cmidrule(lr){3-5}
Precision (QPT) \citep{kumar2024scalingprecision}
& $L=A\left((1-e^{-P/\bar{\gamma}})^3N\right)^{-\alpha}
+BD^{-\beta}+E$
& \xmark & \xmark & \xmark \\
\cmidrule(lr){1-1}\cmidrule(lr){2-2}\cmidrule(lr){3-5}
Compression \citep{frantar2025compression}
& $L=A\left(\rho_N(\psi_f)N\right)^{-\alpha}
+BD^{-\beta}+E$
& \xmark & \xmark & \xmark \\
\cmidrule(lr){1-1}\cmidrule(lr){2-2}\cmidrule(lr){3-5}
QAT \citep{chen2025qatscaling}
& $L=AN^{-\alpha}+BD^{-\beta}+E
+kD^{\gamma_D}N^{-\gamma_N}(\log_2 G)^{\gamma_G}$
& \xmark & \xmark & \xmark \\
\cmidrule(lr){1-1}\cmidrule(lr){2-2}\cmidrule(lr){3-5}
Representations \citep{panferov2026unified}
& $L=A\left(N\,\widetilde{\rho}(\mathrm{GMSE}(\psi_f))\right)^{-\alpha}
+BD^{-\beta}+E$
& \xmark & \xmark & \xmark \\
\cmidrule(lr){1-1}\cmidrule(lr){2-2}\cmidrule(lr){3-5}
Quartet \citep{castro2025quartet}
& $L=\left(
A\left(\rho_N(\psi_f)N\right)^{-\alpha}
+B\left(\rho_D(\psi_b)D\right)^{-\beta}
\right)^\gamma+E$
& \cmark & \xmark & \xmark \\
\cmidrule(lr){1-1}\cmidrule(lr){2-2}\cmidrule(lr){3-5}
{\bf Mixed (QPT)} (Ours)
& \begin{tabular}{@{}l@{}}
(Section~\ref{sec:revised})\;
$\begin{aligned}[c]
L=&\left(
A\left(\rho_N(\psi_f)N\right)^{-\alpha}
+B\left(\rho_D(\psi_b)D\right)^{-\beta}
\right)^\gamma \\
&+\epsilon_f(\psi_f)+\epsilon_b(\psi_b) + E,
\end{aligned}$ 
\vspace{0.5em}
\\
(Section~\ref{sec:mixed})\;
$L=\left(L_2-L_1\right)p^{\gamma_S}+L_1$
\end{tabular}
& \cmark & \cmark & \cmark 
\vspace{0.2em}
\\
\bottomrule
\end{tabular}
\end{adjustbox}
\end{table*}

We first consider mixed precision across the forward and backward passes, where the two passes use quantization schemes $\psi_f$ and $\psi_b$ that may differ, as in the Quartet law \citep{castro2025quartet}, with uniform precision as the special case of matching schemes.
Existing scaling laws for quantized pretraining treat forward quantization as reducing the effective parameter count \citep{kumar2024scalingprecision,frantar2025compression,panferov2026unified} and backward quantization as reducing the effective training token count \citep{castro2025quartet}, implying that quantization penalties can be offset by scaling model size or data (Table~\ref{tab:scaling-law-formulations}). However, quantized training can also introduce optimization error that does not decrease with more data or parameters, as weights can oscillate across rounding boundaries in the forward pass \citep{defossez2021differentiable,nagel2022overcoming} and gradient estimates can be biased in the backward pass \citep{ajalloeian2020convergence}.

We develop a scaling law that accounts for this persistent optimization error,
\begin{equation}
L =
\left(
A\left(\rho_N(\psi_f)N\right)^{-\alpha}
+
B\left(\rho_D(\psi_b)D\right)^{-\beta}
\right)^\gamma
+\epsilon_f(\psi_f)+\epsilon_b(\psi_b)+E,
\label{eq:revised-law}
\end{equation}
where $\rho_N$ and $\rho_D$ rescale $N$ and $D$ according to the forward and backward schemes $\psi_f$ and $\psi_b$, and $\epsilon_f,\epsilon_b\ge0$ capture persistent optimization error introduced by each pass. Section~\ref{sec:law} formalizes this intuition using optimization theory, and Sections~\ref{sec:setup} and \ref{sec:result} empirically validate that the resulting law predicts held-out loss more accurately than existing laws.

\subsection{Scaling law formulation}
\label{sec:law}

Quantization during pretraining affects both the representational capacity of the model and the accuracy of its training updates. To separate these effects, we decompose the loss gap into a capacity gap and an optimization gap \citep{kexuefm-11833}, which guides the formulation of our law. For a language model with parameters $\theta\in\mathbb{R}^N$, let $\mathcal{L}_N(\theta)$ denote the population loss. For a forward quantization scheme $\psi_f$, let $Q(\cdot;\psi_f):\mathbb{R}^N\to\mathbb{R}^N$ denote the corresponding quantization map. We define the optimal loss of the quantized model and the irreducible loss over the model family, respectively, as
\[
\mathcal{L}_{N,\psi_f}^{*}
\coloneqq \inf_{\theta\in\mathbb{R}^N}\mathcal{L}_N(Q(\theta;\psi_f)),
\qquad
E\coloneqq \inf_N\inf_{\theta\in\mathbb{R}^N}\mathcal{L}_N(\theta)
\qquad
(\mathcal{L}_{N,\psi_f}^{*}\ge E).
\]
Consider training for $T$ steps, with weights $\theta_t$ and quantized weights $\hat{\theta}_t=Q(\theta_t;\psi_f)$ at step $t\in[T]$. The gap between the average expected loss of the quantized model and $E$ decomposes as
\[
\mathbb{E}_{J,\theta_J}\!\left[\mathcal{L}_N(\hat{\theta}_J)\right]-E
=
\underbrace{\left(\mathcal{L}_{N,\psi_f}^{*}-E\right)}_{\text{capacity gap}}
+
\underbrace{\left(
\mathbb{E}_{J,\theta_J}\!\left[\mathcal{L}_N(\hat{\theta}_J)\right]
-\mathcal{L}_{N,\psi_f}^{*}
\right)}_{\text{optimization gap}},
\]
where $J$ is sampled uniformly from $[T]$. The first term represents the model \textit{capacity gap} due to the limited number of parameters and the restricted numerical representations imposed by forward quantization. The second term represents the \textit{optimization gap}, measuring how far the average expected loss during training remains above the best loss achievable by the quantized model.

{\bf Capacity gap}. The capacity gap $\mathcal{L}_{N,\psi_f}^{*}-E$ depends only on model size $N$ and the forward quantization $\psi_f$. The backward quantization $\psi_b$ only affects training updates without changing the set of functions the model can represent. We therefore model it through the effective parameter count $\rho_N(\psi_f)N$, as in existing laws \citep{kumar2024scalingprecision,castro2025quartet,panferov2026unified}.

{\bf Optimization gap}. To examine how quantization affects optimization, Remark~\ref{rem:upperbound-convex} below bounds the optimization gap for training with quantized weights and biased backward gradients under suitable regularity conditions. Similar analyses appear in \citet{li2017training, panferov2026unified} for quantized weights and \citet{ajalloeian2020convergence} for biased gradients.


\vspace{1em}

\begin{restatable}{remark}{upperboundconvex}
\label{rem:upperbound-convex}
Consider updates $\theta_{t+1}=\theta_t-\eta\hat g_t$, where
$\hat g_t\in\R^N$ is a minibatch gradient estimate at
$\hat\theta_t=Q(\theta_t;\psi_f)$ 
using backward quantization scheme $\psi_b$.
Define $b_t\coloneqq
\E[\hat g_t\mid\theta_t]-\nabla\Lcal_N(\hat\theta_t)$.
For a minimizer $q^*$ of $\Lcal_N$ over $Q(\R^N;\psi_f)$, set
$R_1(N,\psi_f)\coloneqq\E\|\theta_1-q^*\|^2$.
Assume $\psi_f$ is deterministic, $\Lcal_N$ is differentiable and convex,
and, uniformly over $t,T$,
$\E\|\nabla\Lcal_N(\hat\theta_t)\|^2\le G^2$ and
$\hat g_t$, $b_t$, $\hat\theta_t-\theta_t$,
and $\theta_t-q^*$ have bounded second moments.
For learning rate chosen as $\eta=c/\sqrt T \;(c>0)$,
\[
\begin{aligned}
\E_{J,\theta_J}[\Lcal_N(\hat\theta_J)]
-\Lcal_{N,\psi_f}^*
&\le
\underbrace{
\frac{1}{2\sqrt T}
\left[
\frac{R_1(N,\psi_f)}{c}
+\frac{c}{T}\sum_{t=1}^{T}\E\|\hat g_t\|^2
\right]
}_{\text{(a)}\;O(T^{-1/2})}
\\
&+
\underbrace{
G\left(\frac1T\sum_{t=1}^{T}
\E\|\hat\theta_t-\theta_t\|^2\right)^{1/2}
}_{\text{(b)}\;O(1)}
+
\underbrace{
\frac1T\sum_{t=1}^{T}
\E\!\left[\|b_t\|\|\theta_t-q^*\|\right]
}_{\text{(c)}\;O(1)}.
\end{aligned}
\]
\end{restatable}

Each term in the bound corresponds to a term in the scaling law (Eq.~\ref{eq:revised-law}).
\begin{enumerate}[label=(\alph*), leftmargin=*, itemsep=0pt]
    \item[(a)] The first term decreases as $O(T^{-1/2})$ with additional optimization steps, which correspond to training tokens at a fixed batch size. It motivates the effective data size $\rho_D(\psi_b)D$.
    \item[(b)] The second term depends on the forward quantization error $\|\hat\theta_t-\theta_t\|$, which need not vanish with additional training. Example~\ref{ex:persistent-gap} shows that forward quantization can cause an $\Omega(1)$ optimization gap, motivating the persistent forward error term $\epsilon_f(\psi_f)$. Related oscillations and illustrative examples have been studied by \citet{defossez2021differentiable,nagel2022overcoming}.
    \item[(c)] The third term reflects bias $b_t$ in backward gradient estimates, which can persist despite additional training \citep{ajalloeian2020convergence}, motivating the backward error term $\epsilon_b(\psi_b)$.
\end{enumerate}


\vspace{1em}

\begin{example}[Persistent optimization error from forward quantization]
\label{ex:persistent-gap}
Consider $\Lcal_1(\theta)=(\theta-\tfrac14)^2$ and
$Q(\theta;\psi_f)=\lfloor\theta+\tfrac12\rfloor$.
For $T$ divisible by four, initialize $\theta_1=\tfrac12$ and
choose $\eta=c/\sqrt T$ with $0<c\le\tfrac14$.
Using exact gradients at the quantized weights
$\hat\theta_t=Q(\theta_t;\psi_f)$ gives the updates
$\theta_{t+1}=\theta_t-2\eta(\hat\theta_t-\tfrac14)$.
The quantized iterates repeat the cycle $(1,0,0,0)$.
The optimal quantized weight is $q^*=0$, with
$\Lcal_{1,\psi_f}^*=\tfrac1{16}$.
The optimization gap is
\[
\E_J[\Lcal_1(\hat\theta_J)]-\Lcal_{1,\psi_f}^*
=
\frac14\cdot\frac9{16}
+\frac34\cdot\frac1{16}
-\frac1{16}
=\frac18
=\Omega(1).
\]
Thus, the optimization gap does not vanish as $T\to\infty$.
\end{example}


{\bf Proposed scaling law}. Combining the capacity gap, modeled through $\rho_N(\psi_f)N$, with terms (a), (b), and (c) yields our proposed scaling law (Eq.~\ref{eq:revised-law}), where the persistent contributions are modeled as separate additive terms. While term (a) depends on both quantization schemes, we retain only the backward dependence in $\rho_D$ because including $\psi_f$ did not improve held-out loss prediction (Section~\ref{sec:result}). Table~\ref{tab:scaling-law-formulations} highlights the differences between our law and existing formulations.

\begin{table*}[t]
\caption{Overview of the experimental data grids used to evaluate scaling laws for quantized pretraining. Model sizes $N$ are non-embedding parameter counts. TPP denotes tokens per parameter ($D/N$), where $D$ is the training token count. Token budgets show the range covered by each grid, with detailed settings in Appendix~\ref{sec:app-splits}. Each grid covers all combinations of the listed forward and backward schemes at every evaluated model size and token budget, including the BF16 baseline.}
\label{tab:dataset-overview}
\vspace{0.5em}
\centering
\begin{adjustbox}{max width=\textwidth}
\begin{tabular}{lcllll}
\toprule
Data grid (architecture) & 
\# points
& Model sizes $N$
& Token budgets
& Forward schemes $\psi_f$
& Backward schemes $\psi_b$ \\
\midrule
\multirow{2}{*}{Quartet grid (Llama)}
& \multirow{2}{*}{84}
& 30M, 50M,
& \multirow{2}{*}{25--800 TPP}
& \multirow{2}{*}{BF16, MXFP4 (QuEST)}
& \multirow{2}{*}{BF16, MXFP4 (SR)} \\
& & 100M, 200M & & & \\
\cmidrule(lr){1-1}\cmidrule(lr){2-2}\cmidrule(lr){3-3}\cmidrule(lr){4-4}\cmidrule(lr){5-6}
\multirow{2}{*}{Our grid (Qwen3)}
& \multirow{2}{*}{375}
& 280M, 430M,
& \multirow{2}{*}{12.8B--35.8B}
& BF16, MXFP8,
& BF16, MXFP8, NVFP4 (MS-EDEN), \\
& & 640M, 910M & &
NVFP4 (4/6 + 1D)
& NVFP4 (RHT + SR), NVFP4 (SR) \\
\bottomrule
\end{tabular}
\end{adjustbox}
\end{table*}

\subsection{Experimental setup}
\label{sec:setup}

We evaluate our scaling law on two pretraining grids: (1) the published Quartet experiments with Llama models \citep{castro2025quartet} and (2) our Qwen3 experiments. For each grid, we fit the laws on calibration data and evaluate predictions on held-out runs with larger model sizes or token budgets.


{\bf Data grids}. Table~\ref{tab:dataset-overview} details the two grids, the Quartet grid with 84 published runs of Llama-style models \citep{touvron2023llama} over two forward and two backward schemes \citep{castro2025quartet,panferov2025quest}, and our grid with 375 runs of Qwen3-based models \citep{qwen3} over three forward and five backward schemes selected in Section~\ref{sec:recipes}, covering every forward--backward combination (see Appendix~\ref{sec:app-training} for details).

{\bf Evaluation protocol}. We compare our law with the Quartet law \citep{castro2025quartet} under \textit{model size extrapolation}, fitting on the three smaller models and evaluating on the largest, and \textit{data extrapolation}, fitting on smaller token budgets and evaluating at the largest (Appendix~\ref{sec:app-splits}). Following \citet{castro2025quartet}, we fit each law in two stages, first to BF16 runs and then, with the baseline coefficients frozen, to quantized runs (Appendix~\ref{sec:app-fitting}). We report $R^2$, mean squared error (MSE), and mean absolute percentage error (MAPE) on held-out quantized runs.

\subsection{Experimental results}
\label{sec:result}

\begin{table*}[t]
\caption{Held-out loss prediction under model size and data extrapolation on the Quartet grid (Llama) and our grid (Qwen3). Each law is fitted to the calibration split of each setting and evaluated on the quantized runs in the held-out split. The best results are shown in bold.}
\label{tab:scaling-law-comparison}
\vspace{0.5em}
\centering
\begin{adjustbox}{max width=\textwidth}
\begin{tabular}{ll ccc ccc}
\toprule
& & \multicolumn{3}{c}{Quartet grid (Llama)}
  & \multicolumn{3}{c}{Our grid (Qwen3)} \\
\cmidrule(lr){3-5}\cmidrule(lr){6-8}
Evaluation & Scaling law
& $R^2 \uparrow$ & MSE $\downarrow$ & MAPE (\%) $\downarrow$
& $R^2 \uparrow$ & MSE $\downarrow$ & MAPE (\%) $\downarrow$ \\
\cmidrule(lr){1-1}\cmidrule(lr){2-2}
\cmidrule(lr){3-5}\cmidrule(lr){6-8}
Model size extrapolation & Quartet
& 0.964 & 1.6e-4 & 0.386
& 0.990 & 8.4e-6 & 0.076 \\
& Ours
& \textbf{0.997} & \textbf{1.4e-5} & \textbf{0.096}
& \textbf{0.994} & \textbf{5.3e-6} & \textbf{0.059} \\
\cmidrule(lr){1-1}\cmidrule(lr){2-2}
\cmidrule(lr){3-5}\cmidrule(lr){6-8}
Data extrapolation & Quartet
& 0.995 & 9.0e-5 & 0.234
& 0.992 & 2.7e-5 & 0.121 \\
& Ours
& \textbf{0.997} & \textbf{5.1e-5} & \textbf{0.217}
& \textbf{0.998} & \textbf{6.1e-6} & \textbf{0.064} \\
\bottomrule
\end{tabular}
\end{adjustbox}
\end{table*}


Table~\ref{tab:scaling-law-comparison} reports model size and data extrapolation results on both grids. Our law outperforms the Quartet law across all settings, reducing MSE by 1.6$\times$ to 11$\times$, e.g., from $1.6\times10^{-4}$ to $1.4\times10^{-5}$ (11$\times$) for model size extrapolation on the Quartet grid and from $2.7\times10^{-5}$ to $6.1\times10^{-6}$ (4.4$\times$) for data extrapolation on our grid.
Figures~\ref{fig:quartet-model-extrapolation-calibration} and~\ref{fig:quartet-model-extrapolation} illustrate this on the Quartet grid with MXFP4 QuEST in the forward pass and MXFP4 stochastic rounding in the backward pass, showing the fits on 30M--100M models and the extrapolation to held-out 200M models. Our predictions match more closely the measured losses of the 200M models across token budgets, showing that modeling persistent optimization error improves extrapolation to larger models and token budgets.
Appendix~\ref{sec:app-fits} further provides a comprehensive comparison of both laws across all quantization scheme combinations, grids, and extrapolation settings (Figures~\ref{fig:app-fits-quartet-model}--\ref{fig:app-fits-qwen3-data-ours}).

\begin{wraptable}{r}{0.48\textwidth}
\vspace{-2em}
\centering
\caption{Ablations of our scaling law,
reporting geometric-mean held-out MSE across model
size and data extrapolation on the Quartet and Qwen3 grids. Variants
modify our law as indicated. Best result is bolded.}
\vspace{0.5em}
\label{tab:law-ablations}
\begin{adjustbox}{max width=\linewidth}
\begin{tabular}{lc}
\toprule
Scaling law form & Geo. mean MSE $\downarrow$ \\
\midrule
Quartet & 4.27e-5 \\
\cmidrule(lr){1-1}\cmidrule(lr){2-2}
Ours & \textbf{1.23e-5} \\
- (1) With $\rho_D(\psi_f,\psi_b)$ & 1.29e-5 \\
- (2) Without $\epsilon_b(\psi_b)$ & 1.34e-5 \\
- (3) Without $\epsilon_f(\psi_f)$ & 3.76e-5 \\
\bottomrule
\end{tabular}
\end{adjustbox}
\vspace{-1em}
\end{wraptable}

\paragraph{Ablation results.} We further ablate the design choices in Eq.~\ref{eq:revised-law} by evaluating three variants, (1) allowing $\rho_D$ to depend on both forward and backward quantization schemes, (2) removing the backward error term $\epsilon_b(\psi_b)$, and (3) removing the forward error term $\epsilon_f(\psi_f)$. Table~\ref{tab:law-ablations} reports the geometric mean of held-out MSE across model size and data extrapolation on the Quartet grid and our grid, giving equal weight to relative changes across settings with different error scales. The proposed law achieves the lowest error among the evaluated variants, supporting the chosen formulation.
Appendix~\ref{sec:app-ablation} further evaluates all combinations of these choices, together with adding the error terms inside the outer exponent $\gamma$ instead of outside, and reports the held-out MSE in each setting (Table~\ref{tab:law-ablations-full}).

\begin{figure*}[t]
\centering
\begin{minipage}[t]{0.47\textwidth}
\centering
\QuantizedScalingCalibrationPlot[
  width=0.75\linewidth,
  height=0.45\linewidth,
  legend columns=2,
  legend style={
    at={(0.98,0.98)},
    anchor=north east,
    draw=black!70,
    fill=white,
    line width=0.5pt,
    inner sep=3pt,
    font=\tiny,
    cells={anchor=west},
    row sep=1pt
  }
]
\vspace{-1em}
\caption{Scaling law fits on 30M--100M models (Quartet grid, MXFP4 QuEST forward, MXFP4 stochastic rounding backward).}
\label{fig:quartet-model-extrapolation-calibration}
\end{minipage}
\hfill
\begin{minipage}[t]{0.47\textwidth}
\centering
\QuantizedScalingPlot[
  width=0.75\linewidth,
  height=0.45\linewidth,
  legend columns=2,
  legend style={
    at={(0.98,0.98)},
    anchor=north east,
    draw=black!70,
    fill=white,
    line width=0.5pt,
    inner sep=3pt,
    font=\tiny,
    cells={anchor=west},
    row sep=1pt
  }
]
\vspace{-0.6em}
\caption{Extrapolation to held-out 200M models (Quartet grid, MXFP4 QuEST forward, MXFP4 stochastic rounding backward).}
\label{fig:quartet-model-extrapolation}
\end{minipage}

\end{figure*}

\vspace{-1mm}
\section{Layer-wise mixed-precision scaling law and compute-optimal allocation}
\label{sec:mixed}
\vspace{-1mm}

Practical quantized pretraining recipes often retain higher precision in sensitive layers while using lower-precision computation elsewhere, enabling a flexible trade-off between training loss and cost \citep{nvidia2025nvfp4,li2026nvfp4rl,pan2026snip}. Selecting such configurations before training requires predicting how loss depends not only on the fraction of high-precision blocks, but also on which blocks are assigned higher precision. In this section, we extend Eq.~\ref{eq:revised-law} to model both the fraction of high-precision blocks and how they are selected (Table~\ref{tab:scaling-law-formulations}), and use the extended law to identify compute-optimal precision allocations. Section~\ref{sec:mixed-form} formulates the extension, Section~\ref{sec:mixed-exp} evaluates its predictions, and Section~\ref{sec:compute-optimal} studies compute-optimal precision allocation under a fixed compute budget, showing that the optimal precision increases with the budget.

\vspace{-1mm}
\subsection{Formulation}
\label{sec:mixed-form}
\vspace{-1mm}

Consider a model with $K$ Transformer blocks and two quantization configurations: high precision $\psi_1$ and low precision $\psi_2$, each comprising forward and backward schemes, $\psi_i=(\psi_{f,i},\psi_{b,i})$ for $i\in\{1,2\}$. We interpolate between the two configurations by assigning $\psi_2$ to a subset of the blocks and $\psi_1$ to the rest. A selection strategy $S:[0,1]\to 2^{[K]}$ maps a fraction $p$ to a set of blocks $S(p)\subseteq [K]$ with $|S(p)|=\mathrm{round}(pK)$, and the mixed configuration $\psi^{\mathrm{mix}}_S(p)$ assigns $\psi_2$ to the blocks in $S(p)$ and $\psi_1$ to the remaining blocks. We model its loss as
\begin{equation}
L\big(N,D,\psi^{\mathrm{mix}}_S(p)\big)=\left(L_2-L_1\right)p^{\gamma_S}+L_1,
\label{eq:mixed-law}
\end{equation}
where $L_i=L(N,D,\psi_{f,i},\psi_{b,i})$ is predicted by Eq.~\ref{eq:revised-law}, $L_2\ge L_1$, and $\gamma_S>0$ depends on the strategy $S$. Mixing only the forward or only the backward pass corresponds to $\psi_{b,1}=\psi_{b,2}$ or $\psi_{f,1}=\psi_{f,2}$.

By construction, the loss equals $L_1$ at $p=0$ and $L_2$ at $p=1$, so the strategy only determines how the loss moves between these endpoints. A good strategy assigns $\psi_2$ to the least sensitive blocks first, keeping the loss close to $L_1$ until $p$ becomes large, whereas a poor strategy assigns it to sensitive blocks early and increases the loss quickly. The exponent $\gamma_S$ captures this difference, where a larger $\gamma_S$ indicates that the strategy keeps the sensitive blocks in $\psi_1$ longer, a smaller $\gamma_S$ indicates that it assigns them to $\psi_2$ earlier, and $\gamma_S=1$ corresponds to linear interpolation between $L_1$ and $L_2$.

\vspace{-1mm}
\subsection{Experiments}

\label{sec:mixed-exp}
We evaluate the extended law on Qwen3 models with layer-wise MXFP8/NVFP4 mixed precision. We vary precision in either the forward or backward pass, keeping 25\%, 50\%, or 75\% of the blocks in MXFP8 ($p\in\{0.75, 0.5, 0.25\}$ up to rounding). We compare two strategies for selecting higher-precision blocks: (1) \textit{first/last}, which keeps the first and last blocks in MXFP8, following practical recipes \citep{nvidia2025nvfp4, li2026nvfp4rl}, and (2) \textit{middle}, which keeps the central blocks in MXFP8. We hold out the 910M model, predict endpoints $L_1$ and $L_2$ using laws fitted on smaller models, and fit $\gamma_S$ for each pass and strategy using their mixed-precision runs (see Appendix~\ref{sec:app-mixed} for details).

\begin{figure*}[t]
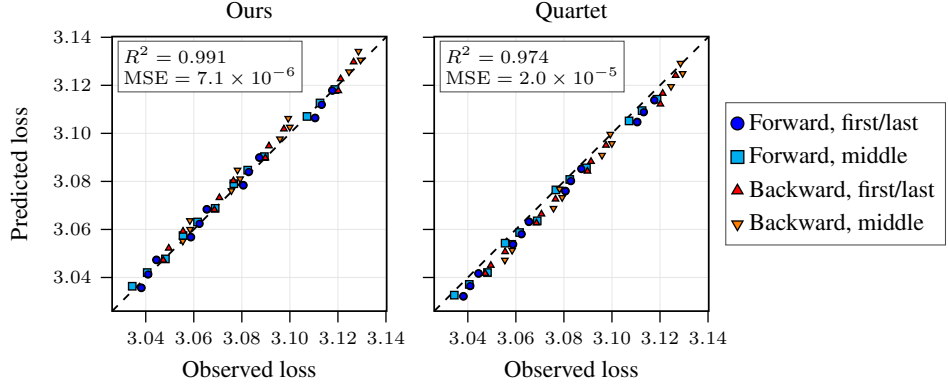

\centering
\MixedParityPlot
\caption{Predicted versus observed loss of layer-wise mixed-precision runs on the held-out 910M model, with $\gamma_S$ fitted on the three smaller models. The endpoints $L_1$ and $L_2$ in Eq.~\ref{eq:mixed-law} are predicted by our law (left) or the Quartet law (right). Markers indicate the mixed pass and the blocks kept in higher precision, and each point corresponds to one fraction $p$ and one token budget.}
\label{fig:mixed-parity}
\end{figure*}

\begin{wraptable}{r}{0.45\linewidth}
\vspace{-2em}
\centering
\caption{Fitted $\gamma_S$ (higher means better strategy) with our law, and the resulting fraction of the loss gap incurred at $p=0.5$.}
\label{tab:mixed-gamma}
\vspace{0.5em}
\begin{adjustbox}{max width=\linewidth}
\begin{tabular}{llcc}
\toprule
Pass & Strategy $S$ & $\gamma_S \uparrow$ & Gap at $p=0.5$ $\downarrow$ \\
\cmidrule(lr){1-1}\cmidrule(lr){2-2}\cmidrule(lr){3-4}
Forward & First/last & 1.10 & 46.7\% \\
\cmidrule(lr){2-2}\cmidrule(lr){3-4}
 & Middle & 1.00 & 50.1\% \\
\cmidrule(lr){1-1}\cmidrule(lr){2-2}\cmidrule(lr){3-4}
Backward & First/last & 1.60 & 33.0\% \\
\cmidrule(lr){2-2}\cmidrule(lr){3-4}
 & Middle & 0.50 & 70.7\% \\
\bottomrule
\vspace{-1em}
\end{tabular}
\end{adjustbox}
\end{wraptable}

\paragraph{Prediction accuracy.} Figure~\ref{fig:mixed-parity} compares the predicted and observed losses of the mixed-precision runs on the held-out 910M model. With endpoints predicted by our law, the extended law closely matches the observed losses across both passes, both strategies, all fractions, and all token budgets ($R^2=0.991$, MSE $=7.1\times10^{-6}$), indicating that $\gamma_S$ fitted on smaller models transfers to a larger model. With endpoints predicted by the Quartet law, the predictions consistently underestimate the observed losses, increasing the MSE by nearly three times ($R^2=0.974$, MSE $=2.0\times10^{-5}$).
This shows that the improved accuracy of our proposed scaling law in Section~\ref{sec:revised} carries over to the layer-wise mixed-precision setting.
Appendix~\ref{sec:app-mixed-fits} further visualizes the fits of Eq.~\ref{eq:mixed-law} for each pass and strategy in terms of the normalized increase in loss (Figure~\ref{fig:app-mixed-normalized}).

\paragraph{First and last blocks matter more in the backward pass.}
Table~\ref{tab:mixed-gamma} reports the fitted $\gamma_S$, where larger values indicate that a strategy retains more sensitive blocks at higher precision and incurs a smaller fraction of the loss gap. Keeping the first and last blocks in higher precision, as proposed in practical NVFP4 training recipes \citep{nvidia2025nvfp4,li2026nvfp4rl}, is particularly effective in the backward pass, as reflected in its higher fitted $\gamma_S$. In the forward pass, both strategies incur about half of the gap at $p=0.5$ (46.7\% and 50.1\%), so placement has little effect. In the backward pass, keeping the first and last blocks in MXFP8 incurs 33.0\% of the gap, compared with 70.7\% for the middle blocks.

\subsection{Compute-optimal precision allocation}
\label{sec:compute-optimal}

Using the scaling laws from Sections~\ref{sec:revised} and \ref{sec:mixed-form}, we identify the compute-optimal forward, backward and layer-wise precision, along with model size and the number of training tokens, under a given compute budget (Figure~\ref{fig:compute-optimal}). Unlike prior precision scaling laws \citep[Section~4.3.2]{kumar2024scalingprecision}, our law predicts that the optimal precision increases with the compute budget.

{\bf Setup}. We consider the setting of mixing (1) MXFP8 and (2) NVFP4 with 4/6 + 1D scaling in the forward pass, and (1) MXFP8 and (2) NVFP4 with MS-EDEN in the backward pass, assigning NVFP4 to fractions $p_f\in[0,1]$ and $p_b\in[0,1]$ of the blocks, respectively, with the \textit{first/last} strategy. Applying Eq.~\ref{eq:mixed-law} to each pass in turn, we predict the loss as
\begin{equation}
L = (1-u)(1-v)\,L_{\mathrm{MX8},\mathrm{MX8}} + u(1-v)\,L_{\mathrm{NV4},\mathrm{MX8}} + (1-u)v\,L_{\mathrm{MX8},\mathrm{NV4}} + uv\,L_{\mathrm{NV4},\mathrm{NV4}},
\label{eq:joint-mixing}
\end{equation}
where $u=p_f^{\gamma_f}$, $v=p_b^{\gamma_b}$, $L_{f,b}$ is the loss predicted by Eq.~\ref{eq:revised-law} with forward format $f$ and backward format $b$, and $\gamma_f$ and $\gamma_b$ are the first/last exponents in Table~\ref{tab:mixed-gamma}. We assume that each exponent is independent of the NVFP4 backward technique and of the format in the other pass, which makes Eq.~\ref{eq:joint-mixing} independent of the order in which the passes are mixed.

{\bf Cost model and optimal allocation}. The forward and backward passes account for roughly one third and two thirds of the $6ND$ training FLOPs, respectively \citep{kaplan2020scaling}. Let $\lambda_{f,\mathrm{MX8}}$, $\lambda_{f,\mathrm{NV4}}$, $\lambda_{b,\mathrm{MX8}}$, and $\lambda_{b,\mathrm{NV4}}$ denote the hardware-dependent speedups of MXFP8 and NVFP4 quantization recipes over BF16 in each pass. Assuming equal cost across blocks and ignoring operations other than matrix multiplications, the training cost in BF16-equivalent FLOPs is
\begin{equation}
C(N,D,p_f,p_b)=2ND\left(\frac{1-p_f}{\lambda_{f,\mathrm{MX8}}}+\frac{p_f}{\lambda_{f,\mathrm{NV4}}}\right)+4ND\left(\frac{1-p_b}{\lambda_{b,\mathrm{MX8}}}+\frac{p_b}{\lambda_{b,\mathrm{NV4}}}\right).
\label{eq:mixed-cost}
\end{equation}
Given a compute budget $C_0$, the \textit{compute-optimal allocation} solves
\begin{equation}
\min_{N,\,D,\,p_f,\,p_b}\; L(N,D,p_f,p_b)
\quad\text{subject to}\quad
C(N,D,p_f,p_b)\le C_0,\quad p_f,p_b\in[0,1],
\label{eq:mixed-opt}
\end{equation}
where $L$ is given by Eq.~\ref{eq:joint-mixing}. We solve this problem via numerical optimization (Appendix~\ref{sec:app-opt}).

Figure~\ref{fig:compute-optimal} shows the solution of Eq.~\ref{eq:mixed-opt}, specifically the optimal model size $N^*$, the optimal number of training tokens $D^*$, and the optimal fractions of blocks in NVFP4 in the forward and backward passes, $p_f^*$ and $p_b^*$, as a function of the compute budget $C_0$ ($x$-axis) and the NVFP4 speedup $\lambda_{\mathrm{NV4}}$ ($y$-axis). We assume that each format has the same speedup in the forward and backward passes, i.e., $\lambda_{f,\mathrm{MX8}}=\lambda_{b,\mathrm{MX8}}=\lambda_{\mathrm{MX8}}$ and $\lambda_{f,\mathrm{NV4}}=\lambda_{b,\mathrm{NV4}}=\lambda_{\mathrm{NV4}}$, and fix $\lambda_{\mathrm{MX8}}=1.8$, following the GEMM kernel speedup reported by NVIDIA Transformer Engine \citep{transformerengine_gemm}. Since actual speedups differ across hardware types and generations, supported data types, and kernel implementations, we report results under other assumed MXFP8 speedups in Appendix~\ref{sec:app-mxfp8-speedup}.

{\bf Optimal precision increases with the budget}. Following a horizontal line in the top panels of Figure~\ref{fig:compute-optimal}, i.e., fixing the NVFP4 speedup $\lambda_{\mathrm{NV4}}$ while increasing the compute budget $C_0$, the optimal NVFP4 fractions $p_f^*$ (top left) and $p_b^*$ (top right) decrease, leaving more blocks in MXFP8.
For example, at $\lambda_{\mathrm{NV4}}=3$, the optimal configuration switches to MXFP8 in the forward pass around $C_0=10^{20}$ FLOPs and in both passes at around $C_0=5\times10^{22}$ FLOPs, with layer-wise mixing smoothing the backward switch into a gradual transition.
This contrasts with prior scaling laws, which predict budget-independent optimal precision when jointly optimizing model size, data, and precision \citep[Section~4.3.2]{kumar2024scalingprecision}, and stems from the persistent optimization error in our law. As $C_0$ grows, model- and data-dependent loss terms shrink while persistent error remains, so the lower error of higher precision eventually outweighs the compute savings of lower precision.

{\bf The backward pass should use low precision first}. Comparing the top panels of Figure~\ref{fig:compute-optimal} at a fixed NVFP4 speedup, the backward pass keeps NVFP4 up to much larger compute budgets than the forward pass. Three factors favor NVFP4 in the backward pass. The backward pass accounts for two thirds of the training FLOPs, so each block in NVFP4 saves more compute. Its larger exponent $\gamma_b$ (Table~\ref{tab:mixed-gamma}) means that assigning blocks to NVFP4 incurs a smaller fraction of the loss gap. Finally, the persistent error of NVFP4 relative to MXFP8 is about half as large in the backward pass as in the forward pass in our fitted law (Appendix~\ref{sec:app-params}). Practitioners with a limited budget for higher precision should therefore allocate it to the forward pass first.


\section{Conclusion}

\label{sec:conclusion}

In this work, we develop a scaling law guided by theoretical analysis to predict the loss of quantized pretraining with mixed precision across the forward and backward passes, modeling persistent optimization error from each pass, and show that it predicts held-out loss more accurately than existing laws. Extending the law to layer-wise mixed precision, we find that keeping the first and last blocks in higher precision is particularly effective in the backward pass. Finally, the law predicts that the compute-optimal precision increases with the compute budget, unlike previous precision scaling laws, since the persistent optimization error does not diminish with scale.

\subsection*{AI Use Statement}

\vspace{-0.5em}

We used large language models and coding agents to assist with drafting and editing the manuscript, implementing plotting and optimization code, and providing assistance with mathematical derivations and proofs. We reviewed and verified the AI-assisted content and take full responsibility for the paper.

\subsection*{Ethics statement}

\vspace{-0.5em}

This work studies scaling laws and optimization for quantized language model pretraining. It does not involve sensitive personal data or real-world deployment. While improved training efficiency may broaden access to capable models, it may also enable harmful downstream uses. We encourage responsible use of these methods.


\subsection*{Reproducibility statement}

\vspace{-0.5em}

We provide the experimental settings, model configurations, quantization schemes, optimization details, and fitting procedures needed to reproduce our results. We also report the scaling-law experiments and ablations used in our analyses.



\subsubsection*{Acknowledgments}
Jinuk Kim is supported by 
Institute of Information \&
Communications Technology Planning \& Evaluation (IITP)
grant funded by the Korea government (MSIT) [No. RS2026-25524173, Ultra-Long-Term Hierarchical Memory
and Reasoning Architecture for Next-Generation Omnimodal Agents, 50\%; and No. RS-2020-II200882, (SW STAR
LAB) Development of deployable learning intelligence via
self-sustainable and trustworthy machine learning, 20\%], and
National Research Foundation of Korea (NRF) grant funded
by the Korea government (MSIT) (No. RS-2024-00354036,
30\%).

\bibliography{iclr2027_conference}
\bibliographystyle{iclr2027_conference}
\clearpage
\appendix
\section{Proof of Remark~\ref{rem:upperbound-convex}}
In this section, we provide the proof of Remark~\ref{rem:upperbound-convex}, restated here for convenience.
\vspace{0.5em}

\upperboundconvex*

\begin{proof}
Write $g_t=\nabla\Lcal_N(\hat\theta_t)$ and
$e_t=\hat\theta_t-\theta_t$.
By convexity and $\E[\hat g_t\mid\theta_t]=g_t+b_t$,
\[
\begin{aligned}
\E[\Lcal_N(\hat\theta_t)-\Lcal_N(q^*)]
&\le \E\langle g_t,\theta_t-q^*+e_t\rangle\\
&=\E\langle\hat g_t,\theta_t-q^*\rangle
-\E\langle b_t,\theta_t-q^*\rangle
+\E\langle g_t,e_t\rangle.
\end{aligned}
\]
The update rule gives
\[
2\eta\langle\hat g_t,\theta_t-q^*\rangle
=\|\theta_t-q^*\|^2-\|\theta_{t+1}-q^*\|^2
+\eta^2\|\hat g_t\|^2.
\]
Summing over $t$ and dropping the nonpositive terminal term yields
\[
\begin{aligned}
\frac1T\sum_{t=1}^T\E[\Lcal_N(\hat\theta_t)-\Lcal_N(q^*)]
&\le \frac{R_1(N,\psi_f)}{2\eta T}
+\frac{\eta}{2T}\sum_{t=1}^T\E\|\hat g_t\|^2\\
&\quad+\frac1T\sum_{t=1}^T\E[\|b_t\|\|\theta_t-q^*\|]
+\frac1T\sum_{t=1}^T\E\langle g_t,e_t\rangle.
\end{aligned}
\]
By Cauchy--Schwarz,
\[
\frac1T\sum_{t=1}^T\E\langle g_t,e_t\rangle
\le
\left(\frac1T\sum_{t=1}^T\E\|g_t\|^2\right)^{1/2}
\left(\frac1T\sum_{t=1}^T\E\|e_t\|^2\right)^{1/2}
\le G\left(\frac1T\sum_{t=1}^T\E\|e_t\|^2\right)^{1/2}.
\]
Taking $J$ uniformly from $\{1,\ldots,T\}$ independently of the
trajectory, and substituting $\eta=c/\sqrt T$ and
$\Lcal_N(q^*)=\Lcal_{N,\psi_f}^*$, proves the claim.
The stated orders follow from the uniform second-moment bounds.
\end{proof}

\section{Implementation details}
\label{sec:imp}

\subsection{Model architecture and training}
\label{sec:app-training}

Our scaling law grid models are derived from the Qwen3-8B architecture \citep{qwen3} by varying the number of Transformer blocks $K$ and setting the remaining dimensions from $K$ following the proportions of Qwen3-8B, which has $K=36$ blocks, width $d_{\mathrm{model}}=4096$, and 32 query heads. The width scales linearly with depth, $d_{\mathrm{model}}=\mathrm{round}(4096K/36,\,128)$, and the number of query heads scales with width, $n_{\mathrm{heads}}=\mathrm{round}(32\,d_{\mathrm{model}}/4096,\,8)$, where $\mathrm{round}(x,m)$ rounds $x$ to the nearest multiple of $m$. All other choices follow Qwen3-8B, including grouped-query attention with 8 key-value heads and a head dimension of 128, an FFN dimension of $3d_{\mathrm{model}}$, QK normalization, and tied input and output embeddings over the Qwen3 vocabulary of 151,936 tokens. Table~\ref{tab:model-ladder} lists the resulting configurations, where non-embedding parameter counts include normalization weights.

\begin{table}[t]
\caption{Model configurations derived from Qwen3-8B. $K$ denotes the number of Transformer blocks. Parameter counts are given without and with the tied embeddings.}
\label{tab:model-ladder}
\centering
\begin{adjustbox}{max width=\linewidth}
\begin{tabular}{lccccccc}
\toprule
Model & $K$ & $d_{\mathrm{model}}$ & Query heads & KV heads & FFN dim. & Non-emb. params & Total params \\
\midrule
280M & 12 & 1408 & 8  & 8 & 4224  & 283M  & 497M \\
430M & 14 & 1536 & 16 & 8 & 4608  & 429M  & 663M \\
640M & 16 & 1792 & 16 & 8 & 5376  & 639M  & 911M \\
910M & 18 & 2048 & 16 & 8 & 6144  & 906M  & 1217M \\
\midrule
Qwen3-8B & 36 & 4096 & 32 & 8 & 12288 & 6946M & 7568M \\
\bottomrule
\end{tabular}
\end{adjustbox}
\end{table}

\paragraph{Parameterization and initialization}
We use the CompleteP parameterization \citep{dey2026don} with Qwen3-8B as the base shape, so that hyperparameters tuned on one model transfer across width and depth. For each model size, all runs start from the same initial checkpoint and see the same data order, so BF16 and all quantized recipes differ only in their quantization schemes.

\paragraph{Optimization and data}
We use AdamW \citep{loshchilov2019adamw} ($\beta=(0.9,0.95)$, $\epsilon=10^{-8}$, weight decay $0.1$, gradient clipping $1.0$) with batches of 1024 sequences of length 4096 (about 4.2M tokens), and a warmup--stable--decay schedule \citep{hu2024minicpm} with 244 warmup steps and a linear decay to 1\% of the peak over the final 10\% of steps. A BF16 sweep on the 280M model at 5.7B tokens (20 tokens per parameter) gives an optimal peak learning rate of $1.2\times10^{-3}$, which is also optimal for the 170M and 430M models, confirming transfer across width and depth \citep{dey2026don}. Scaling it to the smallest token budget of each model (Table~\ref{tab:evaluation-settings}) with $\eta\propto D^{-0.32}$ \citep{bjorck2024scaling} gives $9.25\times10^{-4}$ for the three smaller models and $8.27\times10^{-4}$ for the 910M model, which we use for all budgets and quantization schemes.
We report validation loss on C4 \citep{dodge2021documenting}.

\paragraph{Token budgets}
We train every model with at least 20 tokens per parameter \citep{hoffmann2022training}, using the token budgets listed in Table~\ref{tab:evaluation-settings}. The three smaller models share the budgets of the 640M model, ranging from 12.8B to 35.8B tokens, and the 910M model uses budgets from 18.1B to 29.9B tokens. For cost efficiency, we obtain each token budget by resuming from the run of a smaller budget before its decay phase and applying its own 10\% decay.

\subsection{Evaluation splits}
\label{sec:app-splits}
Table~\ref{tab:evaluation-settings} lists the calibration and held-out splits for the two extrapolation settings in Section~\ref{sec:setup}. In model size extrapolation, each law is fitted to runs with the three smaller models of each grid and evaluated on the largest model, which is 200M for the Quartet grid and 910M for our grid. In data extrapolation, each law is fitted to runs with the three smaller models at all but the largest token budget and evaluated at that budget, which is 800 tokens per parameter for the Quartet grid and 35.8B tokens for our grid. The largest model is therefore excluded from data extrapolation. In both settings, the laws are fitted only to the calibration split.

\begin{table*}[t]
\caption{Calibration and held-out splits for model size and data extrapolation. \# points denotes the number of data points in each split, including all forward and backward scheme combinations. The Quartet grid uses tokens-per-parameter ratios ($D/N$), whereas our grid uses training token counts ($D$). $N$ denotes the non-embedding parameter count.}
\label{tab:evaluation-settings}
\vspace{0.5em}
\centering
\begin{adjustbox}{max width=\textwidth}
\begin{tabular}{llclcl}
\toprule
& & \multicolumn{2}{c}{Calibration}
  & \multicolumn{2}{c}{Held-out} \\
\cmidrule(lr){3-4}\cmidrule(lr){5-6}
Data grid & Evaluation
& \# points & Model sizes and token budgets
& \# points & Model sizes and token budgets \\
\cmidrule(lr){1-2}\cmidrule(lr){3-4}\cmidrule(lr){5-6}
\multirow{4}{*}{\begin{tabular}{@{}l@{}}Quartet grid\\(Llama)\end{tabular}}
& Model size
& \multirow{2}{*}{72} & $N\in\{30,50,100\}\mathrm{M}$
& \multirow{2}{*}{12} & $N=200\mathrm{M}$ \\
& extrapolation
& & $D/N\in\{25,50,100,200,400,800\}$
& & $D/N\in\{25,50,100\}$ \\
\cmidrule(lr){2-2}\cmidrule(lr){3-4}\cmidrule(lr){5-6}
 & \multirow{2}{*}{Data extrapolation}
& \multirow{2}{*}{60} & $N\in\{30,50,100\}\mathrm{M}$
& \multirow{2}{*}{12} & $N\in\{30,50,100\}\mathrm{M}$ \\
&
& & $D/N\in\{25,50,100,200,400\}$
& & $D/N=800$ \\
\cmidrule(lr){1-2}\cmidrule(lr){3-4}\cmidrule(lr){5-6}
\multirow{4}{*}{\begin{tabular}{@{}l@{}}Our grid\\(Qwen3)\end{tabular}}
& Model size
& \multirow{2}{*}{315} & $N\in\{280,430,640\}\mathrm{M}$
& \multirow{2}{*}{60} & $N=910\mathrm{M}$ \\
& extrapolation
& & $D\in\{12.8,15.3,17.9,21.1,25.5,30.0,35.8\}\mathrm{B}$
& & $D\in\{18.1,21.7,25.4,29.9\}\mathrm{B}$ \\
\cmidrule(lr){2-2}\cmidrule(lr){3-4}\cmidrule(lr){5-6} & \multirow{2}{*}{Data extrapolation}
& \multirow{2}{*}{270} & $N\in\{280,430,640\}\mathrm{M}$
& \multirow{2}{*}{45} & $N\in\{280,430,640\}\mathrm{M}$ \\
&
& & $D\in\{12.8,15.3,17.9,21.1,25.5,30.0\}\mathrm{B}$
& & $D=35.8\mathrm{B}$ \\
\bottomrule
\end{tabular}
\end{adjustbox}
\end{table*}

\subsection{Fitting the scaling laws}
\label{sec:app-fitting}

We fit both our law and the Quartet law in two stages, following \citet{castro2025quartet}, and both stages minimize a Huber loss with threshold $10^{-3}$ on log-loss residuals $\log\hat{L}-\log L$. We use Levenberg--Marquardt with 200 initializations for the baseline fit and a bounded trust-region reflective solver with 20 initializations for the quantization parameters. We repeat both stages independently for each data grid and extrapolation setting using only calibration data.

\subsection{Layer-wise mixed-precision experiments}
\label{sec:app-mixed}
We use four Qwen3 model sizes with 280M, 430M, 640M, and 910M non-embedding parameters, each trained with four token budgets, yielding 16 combinations of $N$ and $D$. The three smaller models are trained on 12.8B, 15.3B, 17.9B, and 21.1B tokens, and the largest model on 18.1B, 21.7B, 25.4B, and 29.9B tokens. We mix precision in one pass at a time. For each pass and strategy, we keep 25\%, 50\%, or 75\% of the blocks in MXFP8, with each selection containing the smaller ones, so that $p=1-k/K$ for $k$ MXFP8 blocks out of $K$ equals 0.75, 0.5, or 0.25 up to rounding.
\begin{itemize} \itemsep=0pt
    \item \textit{Forward mixing.} $\psi_1$ and $\psi_2$ use MXFP8 and NVFP4 with 4/6 + 1D scaling in the forward pass, respectively, and both use MS-EDEN in the backward pass.
    \item \textit{Backward mixing.} $\psi_1$ and $\psi_2$ use MXFP8 with round-to-nearest and NVFP4 with SR in the backward pass, respectively, and both use 4/6 + 1D scaling in the forward pass.
\end{itemize}
We evaluate the extended law under model size extrapolation, holding out the 910M model. To predict the endpoints $L_1$ and $L_2$, we use each scaling law fitted to the calibration split for model size extrapolation on our grid (Appendix~\ref{sec:app-splits}), which excludes the 910M model. 
We then fit $\gamma_S$ in Eq.~\ref{eq:mixed-law} for each pass and strategy to all mixed-precision runs on the three smaller models, using predicted endpoints, so that the 910M model is excluded from both fits.

\section{Ablation over all combinations of design choices}
\label{sec:app-ablation}

\begin{table*}[t]
\caption{Ablation of our scaling law over all combinations of its design choices. We report held-out MSE under model size and data extrapolation on both grids and their geometric mean. Variants (1)--(3) are those in Table~\ref{tab:law-ablations}, and -- marks the remaining combinations. The best geometric mean is shown in bold.}
\label{tab:law-ablations-full}
\vspace{0.5em}
\centering
\begin{adjustbox}{max width=\textwidth}
\begin{tabular}{c cccc cc cc c}
\toprule
& \multicolumn{4}{c}{Form}
& \multicolumn{2}{c}{Quartet grid (Llama)}
& \multicolumn{2}{c}{Our grid (Qwen3)} & \\
\cmidrule(lr){2-5}\cmidrule(lr){6-7}\cmidrule(lr){8-9}
Scaling law & $\epsilon_f$ & $\epsilon_b$ & $\rho_D$ & $\epsilon$ placement
& Model size & Data
& Model size & Data
& Geo. mean $\downarrow$ \\
\cmidrule(lr){1-1}\cmidrule(lr){2-5}\cmidrule(lr){6-7}\cmidrule(lr){8-9}\cmidrule(lr){10-10}
-- & \xmark & \xmark & $\psi_f,\psi_b$ & -- & 1.27e-4 & 7.98e-5 & 7.28e-6 & 2.61e-5 & 3.72e-5 \\
Quartet & \xmark & \xmark & $\psi_b$ & -- & 1.65e-4 & 9.02e-5 & 8.39e-6 & 2.66e-5 & 4.27e-5 \\
\cmidrule(lr){1-1}\cmidrule(lr){2-5}\cmidrule(lr){6-7}\cmidrule(lr){8-9}\cmidrule(lr){10-10}
-- & \cmark & \xmark & $\psi_f,\psi_b$ & Inside & 2.74e-5 & 5.63e-5 & 4.47e-6 & 2.12e-5 & 1.96e-5 \\
-- & \cmark & \xmark & $\psi_b$ & Inside & 2.87e-5 & 7.10e-5 & 5.17e-6 & 2.31e-5 & 2.22e-5 \\
-- & \cmark & \xmark & $\psi_f,\psi_b$ & Outside & 1.95e-5 & 2.02e-5 & 3.69e-6 & 2.20e-5 & 1.34e-5 \\
Variant (2) & \cmark & \xmark & $\psi_b$ & Outside & 1.67e-5 & 1.86e-5 & 4.32e-6 & 2.38e-5 & 1.34e-5 \\
\cmidrule(lr){1-1}\cmidrule(lr){2-5}\cmidrule(lr){6-7}\cmidrule(lr){8-9}\cmidrule(lr){10-10}
-- & \xmark & \cmark & $\psi_f,\psi_b$ & Inside & 9.97e-5 & 1.02e-4 & 8.19e-6 & 4.98e-6 & 2.54e-5 \\
-- & \xmark & \cmark & $\psi_b$ & Inside & 1.76e-4 & 1.37e-4 & 1.06e-5 & 8.08e-6 & 3.79e-5 \\
-- & \xmark & \cmark & $\psi_f,\psi_b$ & Outside & 1.15e-4 & 1.05e-4 & 6.07e-6 & 6.14e-6 & 2.59e-5 \\
Variant (3) & \xmark & \cmark & $\psi_b$ & Outside & 1.84e-4 & 1.35e-4 & 8.82e-6 & 9.12e-6 & 3.76e-5 \\
\cmidrule(lr){1-1}\cmidrule(lr){2-5}\cmidrule(lr){6-7}\cmidrule(lr){8-9}\cmidrule(lr){10-10}
-- & \cmark & \cmark & $\psi_f,\psi_b$ & Inside & 3.32e-5 & 7.35e-5 & 7.74e-6 & 3.98e-6 & 1.66e-5 \\
-- & \cmark & \cmark & $\psi_b$ & Inside & 3.86e-5 & 9.75e-5 & 8.17e-6 & 5.43e-6 & 2.02e-5 \\
Variant (1) & \cmark & \cmark & $\psi_f,\psi_b$ & Outside & 2.14e-5 & 5.24e-5 & 4.81e-6 & 5.12e-6 & 1.29e-5 \\
\textbf{Ours} & \cmark & \cmark & $\psi_b$ & Outside & 1.43e-5 & 5.06e-5 & 5.29e-6 & 6.07e-6 & \textbf{1.23e-5} \\
\bottomrule
\end{tabular}
\end{adjustbox}
\end{table*}

While Table~\ref{tab:law-ablations} varies one design choice of our law at a time, Table~\ref{tab:law-ablations-full} evaluates all combinations of three choices, (i) which persistent error terms are included among $\epsilon_f(\psi_f)$ and $\epsilon_b(\psi_b)$, (ii) whether $\rho_D$ depends on the backward scheme $\psi_b$ only or on both schemes $\psi_f$ and $\psi_b$, and (iii) whether the error terms are added outside $\gamma$, as in Eq.~\ref{eq:revised-law}, or inside it, i.e., $L=\big(A(\rho_N N)^{-\alpha}+B(\rho_D D)^{-\beta}+\epsilon_f+\epsilon_b\big)^{\gamma}+E$. Each of the resulting 14 forms is fitted and evaluated with the same protocol as our law (Section~\ref{sec:setup}). Overall, our law achieves the lowest geometric-mean MSE. 
For (i) the persistent error terms, removing $\epsilon_f$ increases the geometric-mean MSE by 1.53$\times$ to 3.04$\times$ (rows 7--10 vs.\ rows 11--14), and removing $\epsilon_b$ increases it by 1.04$\times$ to 1.18$\times$ (rows 3--6 vs.\ rows 11--14).
For (ii) the dependence of $\rho_D$, the effect of a joint $\rho_D(\psi_f,\psi_b)$ is mixed across the other choices, and it does not improve on our law (row 13 vs.\ row 14).
For (iii) the placement of the error terms, placing them inside $\gamma$ increases the geometric-mean MSE by 1.31$\times$ on average.

\section{Fitted scaling law parameters}
\label{sec:app-params}

Table~\ref{tab:fitted-params} lists the parameters of our scaling law (Eq.~\ref{eq:revised-law}) fitted to the calibration split of each grid and extrapolation setting. The shared coefficients $E$, $A$, $B$, $\alpha$, $\beta$, and $\gamma$ are fitted in the first stage to the BF16 runs, and the scheme-dependent parameters $\rho_N$, $\rho_D$, $\epsilon_f$, and $\epsilon_b$ are fitted in the second stage to the quantized runs, with $\rho_N=\rho_D=1$ and $\epsilon_f=\epsilon_b=0$ for BF16 by construction. $N$ and $D$ are measured in non-embedding parameters and tokens, respectively.

\begin{table*}[t]
\caption{Fitted parameters of our scaling law on the Quartet grid (Llama) and our grid (Qwen3), for each extrapolation setting (model size and data). The shared coefficients are fitted to the BF16 runs, and the scheme-dependent parameters to the quantized runs, with $\rho_N=\rho_D=1$ and $\epsilon_f=\epsilon_b=0$ for BF16 by construction. $N$ and $D$ are measured in non-embedding parameters and tokens.}
\label{tab:fitted-params}
\vspace{0.5em}
\centering
\begin{adjustbox}{max width=\textwidth}
\begin{tabular}{ll cc cc}
\toprule
& & \multicolumn{2}{c}{Quartet grid (Llama)}
  & \multicolumn{2}{c}{Our grid (Qwen3)} \\
\cmidrule(lr){3-4}\cmidrule(lr){5-6}
Parameter & Scheme
& Model size & Data
& Model size & Data \\
\cmidrule(lr){1-1}\cmidrule(lr){2-2}
\cmidrule(lr){3-4}\cmidrule(lr){5-6}
$E$ & \multicolumn{1}{c}{--} & $1.650$ & $1.763$ & $2.546$ & $2.574$ \\
$A$ & \multicolumn{1}{c}{--} & $1.10\times10^{5}$ & $3.50\times10^{5}$ & $7.62\times10^{4}$ & $9.38\times10^{4}$ \\
$B$ & \multicolumn{1}{c}{--} & $2.50\times10^{5}$ & $6.70\times10^{5}$ & $8.21\times10^{5}$ & $1.04\times10^{6}$ \\
$\alpha$ & \multicolumn{1}{c}{--} & $0.622$ & $0.701$ & $0.629$ & $0.642$ \\
$\beta$ & \multicolumn{1}{c}{--} & $0.550$ & $0.605$ & $0.631$ & $0.642$ \\
$\gamma$ & \multicolumn{1}{c}{--} & $0.322$ & $0.315$ & $0.767$ & $0.787$ \\
\cmidrule(lr){1-1}\cmidrule(lr){2-2}
\cmidrule(lr){3-4}\cmidrule(lr){5-6}
$\rho_N$ & MXFP4 (QuEST) & $0.770$ & $0.798$ & -- & -- \\
& MXFP8 & -- & -- & $0.987$ & $0.984$ \\
& NVFP4 (4/6 + 1D) & -- & -- & $0.922$ & $0.923$ \\
\cmidrule(lr){1-1}\cmidrule(lr){2-2}
\cmidrule(lr){3-4}\cmidrule(lr){5-6}
$\rho_D$ & MXFP4 (SR) & $0.951$ & $0.973$ & -- & -- \\
& MXFP8 & -- & -- & $0.998$ & $1.000$ \\
& NVFP4 (MS-EDEN) & -- & -- & $1.000$ & $1.000$ \\
& NVFP4 (RHT + SR) & -- & -- & $1.000$ & $1.000$ \\
& NVFP4 (SR) & -- & -- & $1.000$ & $1.000$ \\
\cmidrule(lr){1-1}\cmidrule(lr){2-2}
\cmidrule(lr){3-4}\cmidrule(lr){5-6}
$\epsilon_f$ & MXFP4 (QuEST) & $0.0340$ & $0.0397$ & -- & -- \\
& MXFP8 & -- & -- & $0.0021$ & $0.0016$ \\
& NVFP4 (4/6 + 1D) & -- & -- & $0.0151$ & $0.0152$ \\
\cmidrule(lr){1-1}\cmidrule(lr){2-2}
\cmidrule(lr){3-4}\cmidrule(lr){5-6}
$\epsilon_b$ & MXFP4 (SR) & $0.0076$ & $0.0100$ & -- & -- \\
& MXFP8 & -- & -- & $0.0004$ & $0.0008$ \\
& NVFP4 (MS-EDEN) & -- & -- & $0.0076$ & $0.0076$ \\
& NVFP4 (RHT + SR) & -- & -- & $0.0076$ & $0.0076$ \\
& NVFP4 (SR) & -- & -- & $0.0215$ & $0.0211$ \\
\bottomrule
\end{tabular}
\end{adjustbox}
\end{table*}

\section{Scaling law fits for all schemes}
\label{sec:app-fits}

\begin{figure}[t]
\centering
\includegraphics[width=\linewidth]{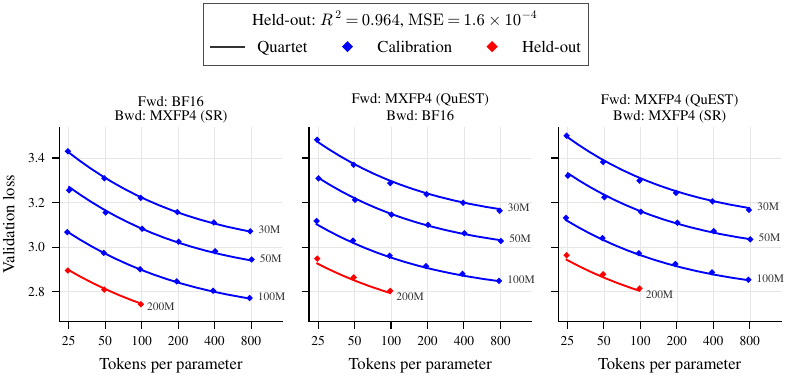}\\[0.2em]
{\small (a) Quartet law}\\[0.8em]
\includegraphics[width=\linewidth]{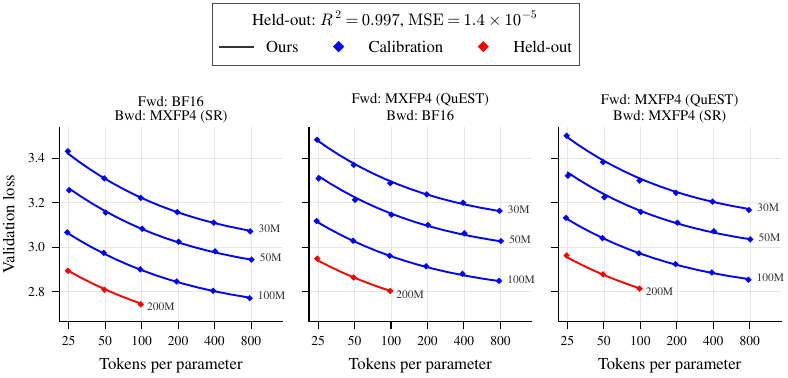}\\[0.2em]
{\small (b) Our law}
\caption{Quartet grid (Llama), model size extrapolation, for (a) the Quartet law and (b) our law. Each law is fitted to the 30M--100M models and evaluated on the held-out 200M models.}
\label{fig:app-fits-quartet-model}
\end{figure}

\begin{figure}[t]
\centering
\includegraphics[width=\linewidth]{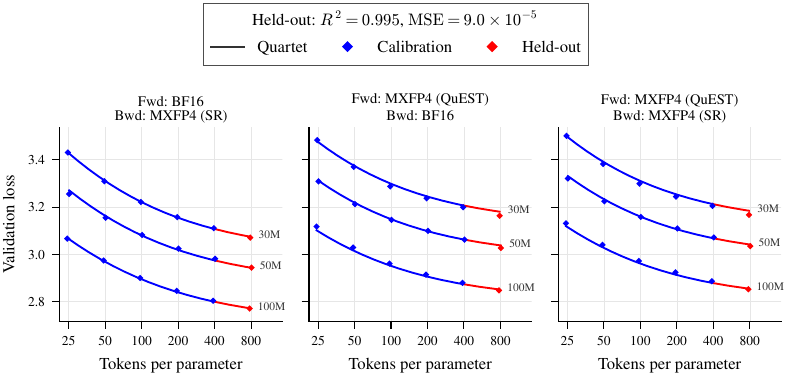}\\[0.2em]
{\small (a) Quartet law}\\[0.8em]
\includegraphics[width=\linewidth]{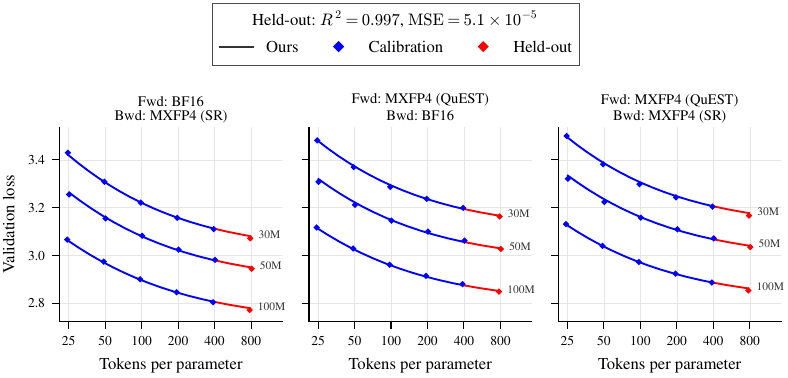}\\[0.2em]
{\small (b) Our law}
\caption{Quartet grid (Llama), data extrapolation, for (a) the Quartet law and (b) our law. Each law is fitted to 25--400 tokens per parameter and evaluated at the held-out 800 tokens per parameter.}
\label{fig:app-fits-quartet-data}
\end{figure}

\begin{figure}[p]
\centering
\includegraphics[width=\linewidth]{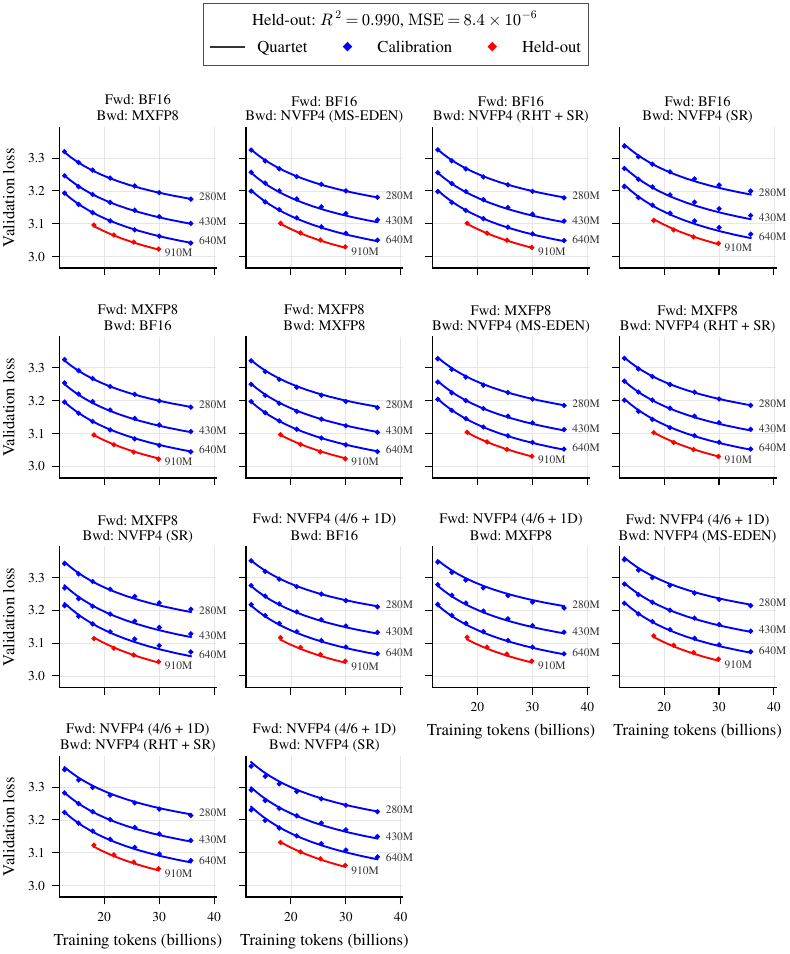}
\caption{Our grid (Qwen3), model size extrapolation, for the Quartet law. The law is fitted to the 280M--640M models and evaluated on the held-out 910M models.}
\label{fig:app-fits-qwen3-model-quartet}
\end{figure}

\begin{figure}[p]
\centering
\includegraphics[width=\linewidth]{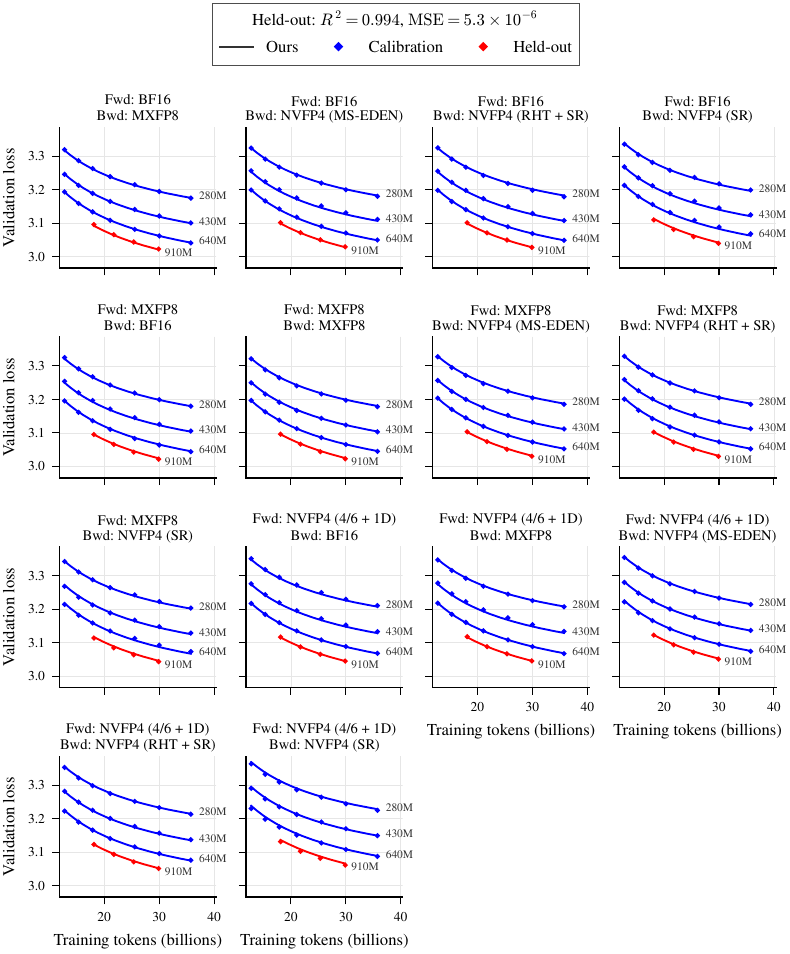}
\caption{Our grid (Qwen3), model size extrapolation, for our law. The law is fitted to the 280M--640M models and evaluated on the held-out 910M models.}
\label{fig:app-fits-qwen3-model-ours}
\end{figure}

\begin{figure}[p]
\centering
\includegraphics[width=\linewidth]{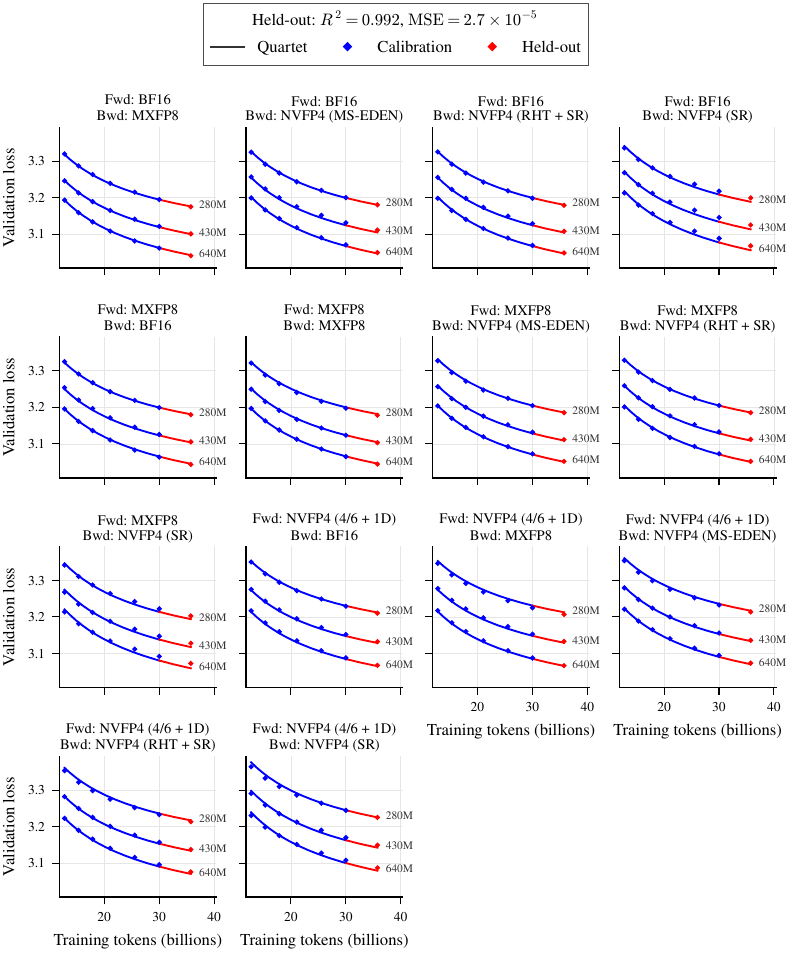}
\caption{Our grid (Qwen3), data extrapolation, for the Quartet law. The law is fitted to all but the largest token budget and evaluated at the held-out 35.8B tokens.}
\label{fig:app-fits-qwen3-data-quartet}
\end{figure}

\begin{figure}[p]
\centering
\includegraphics[width=\linewidth]{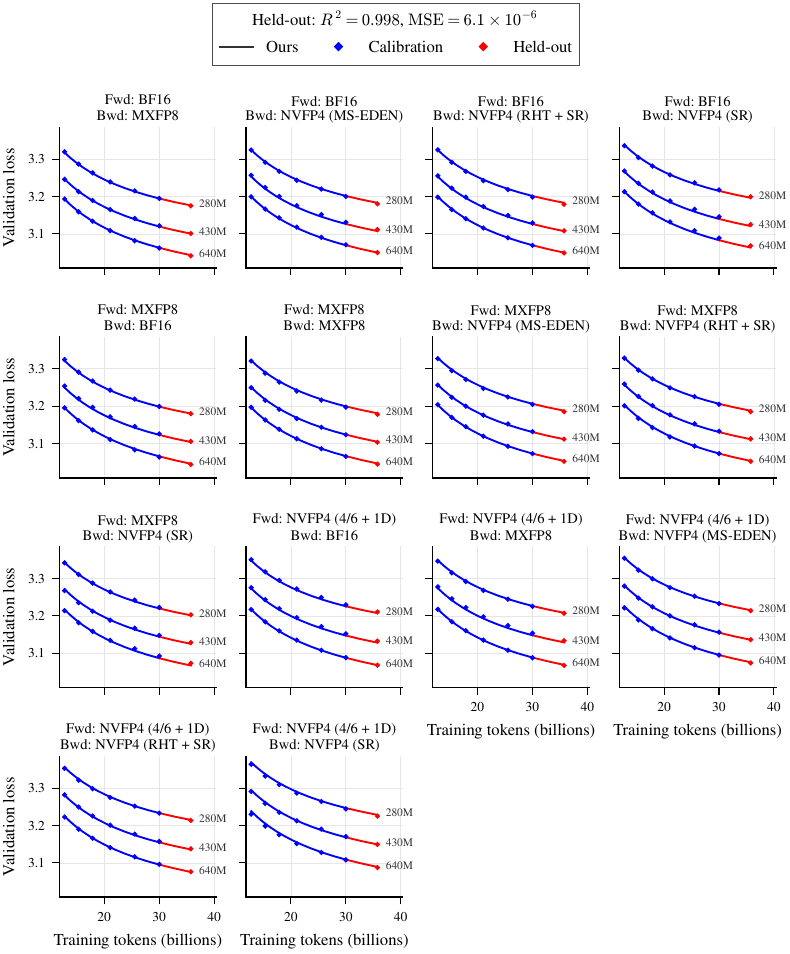}
\caption{Our grid (Qwen3), data extrapolation, for our law. The law is fitted to all but the largest token budget and evaluated at the held-out 35.8B tokens.}
\label{fig:app-fits-qwen3-data-ours}
\end{figure}

Figures~\ref{fig:app-fits-quartet-model}--\ref{fig:app-fits-qwen3-data-ours} show the fits of the Quartet law and our law for all forward and backward scheme combinations on both grids, under model size and data extrapolation. Blue points and curves show the calibration runs and the fitted law, and red points and curves show the held-out runs and the extrapolated predictions. Each legend reports the held-out $R^2$ and MSE in Table~\ref{tab:scaling-law-comparison}.
These figures compare the two laws for each scheme combination on each grid and extrapolation setting.
For instance, on the Quartet grid, the Quartet law underestimates the loss of the held-out 200M models with MXFP4 (QuEST) in the forward pass (Figure~\ref{fig:app-fits-quartet-model}a, middle and right), whereas our law matches it closely (Figure~\ref{fig:app-fits-quartet-model}b).
Similarly, on our grid, the Quartet law underestimates the held-out loss at 35.8B tokens for the runs with NVFP4 (SR) in the backward pass and deviates from their calibration points (Figure~\ref{fig:app-fits-qwen3-data-quartet}, first row, fourth column, and third row, first column), whereas our law matches both closely (Figure~\ref{fig:app-fits-qwen3-data-ours}).

\clearpage

\section{Layer-wise mixed-precision fits}
\label{sec:app-mixed-fits}

\begin{figure}[t]
\centering
\includegraphics[width=\linewidth]{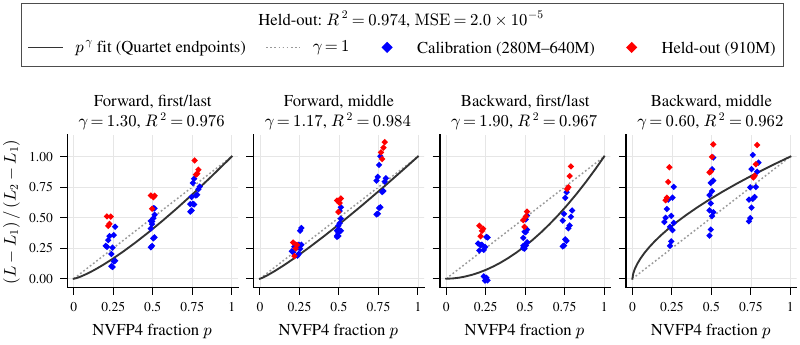}\\[0.2em]
{\small (a) Quartet law endpoints}\\[0.8em]
\includegraphics[width=\linewidth]{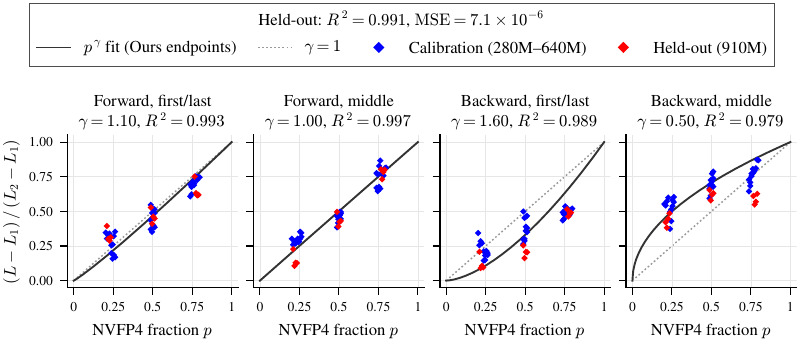}\\[0.2em]
{\small (b) Our law endpoints}
\caption{Normalized increase in loss of layer-wise mixed-precision runs versus the fraction $p$ of blocks in NVFP4, with endpoints $L_1$ and $L_2$ predicted by (a) the Quartet law and (b) our law. Solid curves show $p^{\gamma_S}$ with $\gamma_S$ fitted on the calibration runs, and dotted lines show linear interpolation ($\gamma_S=1$).}
\label{fig:app-mixed-normalized}
\end{figure}

Figure~\ref{fig:app-mixed-normalized} plots the normalized increase in loss $(L-L_1)/(L_2-L_1)$ of each mixed-precision run against the fraction $p$ of blocks in NVFP4, along with the fitted prediction $p^{\gamma_S}$ of Eq.~\ref{eq:mixed-law}, using the endpoints $L_1$ and $L_2$ predicted by the Quartet law and by our law. Blue points show the calibration runs used to fit $\gamma_S$, and red points show the runs on the held-out 910M model. Each legend reports the held-out $R^2$ and MSE in Figure~\ref{fig:mixed-parity}, and each panel title reports the fitted $\gamma_S$ and the held-out $R^2$ for that pass and strategy. With endpoints predicted by our law, the fitted predictions match the normalized losses more closely than with endpoints predicted by the Quartet law, with held-out $R^2$ of 0.979--0.997 versus 0.962--0.984 across panels.

\section{Solving the compute-optimal allocation}
\label{sec:app-opt}

The mixed loss in Eq.~\ref{eq:joint-mixing} combines the four uniform endpoint losses $L_{f,b}(N,D)$ from Eq.~\ref{eq:revised-law}, all evaluated at the same $N$ and $D$, since a mixed configuration trains a single model on a single dataset. We write the cost in Eq.~\ref{eq:mixed-cost} as $C=ND\,c(p_f,p_b)$, where
\begin{equation}
c(p_f,p_b)=2\left(\frac{1-p_f}{\lambda_{f,\mathrm{MX8}}}+\frac{p_f}{\lambda_{f,\mathrm{NV4}}}\right)+4\left(\frac{1-p_b}{\lambda_{b,\mathrm{MX8}}}+\frac{p_b}{\lambda_{b,\mathrm{NV4}}}\right)
\end{equation}
is the training cost per parameter per token, which equals $6$ for BF16. Since the loss decreases in $D$, the budget constraint is tight at the optimum, so $D=C_0/(N\,c(p_f,p_b))$, and we solve
\begin{equation}
\min_{p_f,\,p_b\in[0,1],\;N>0}\; L\!\left(N,\,\frac{C_0}{N\,c(p_f,p_b)},\,p_f,\,p_b\right).
\label{eq:app-opt}
\end{equation}
This problem does not admit a closed-form solution, since the endpoint losses have different optimal splits between parameters and data. We therefore solve it numerically, similar to \citet[Appendix~E.3]{kumar2024scalingprecision}, who solve a related compute-optimal problem via grid search. Specifically, we perform a grid search over $(p_f,p_b,N)$ and refine the best candidates with L-BFGS-B \citep{byrd1995limited}, running it from multiple starting points and keeping the best solution, as in the fitting procedure of \citet[Section~3.3]{hoffmann2022training}. 

\section{Compute-optimal allocation under other MXFP8 speedups}
\label{sec:app-mxfp8-speedup}

\begin{figure}[p]
\centering
\includegraphics[width=0.7\textwidth]{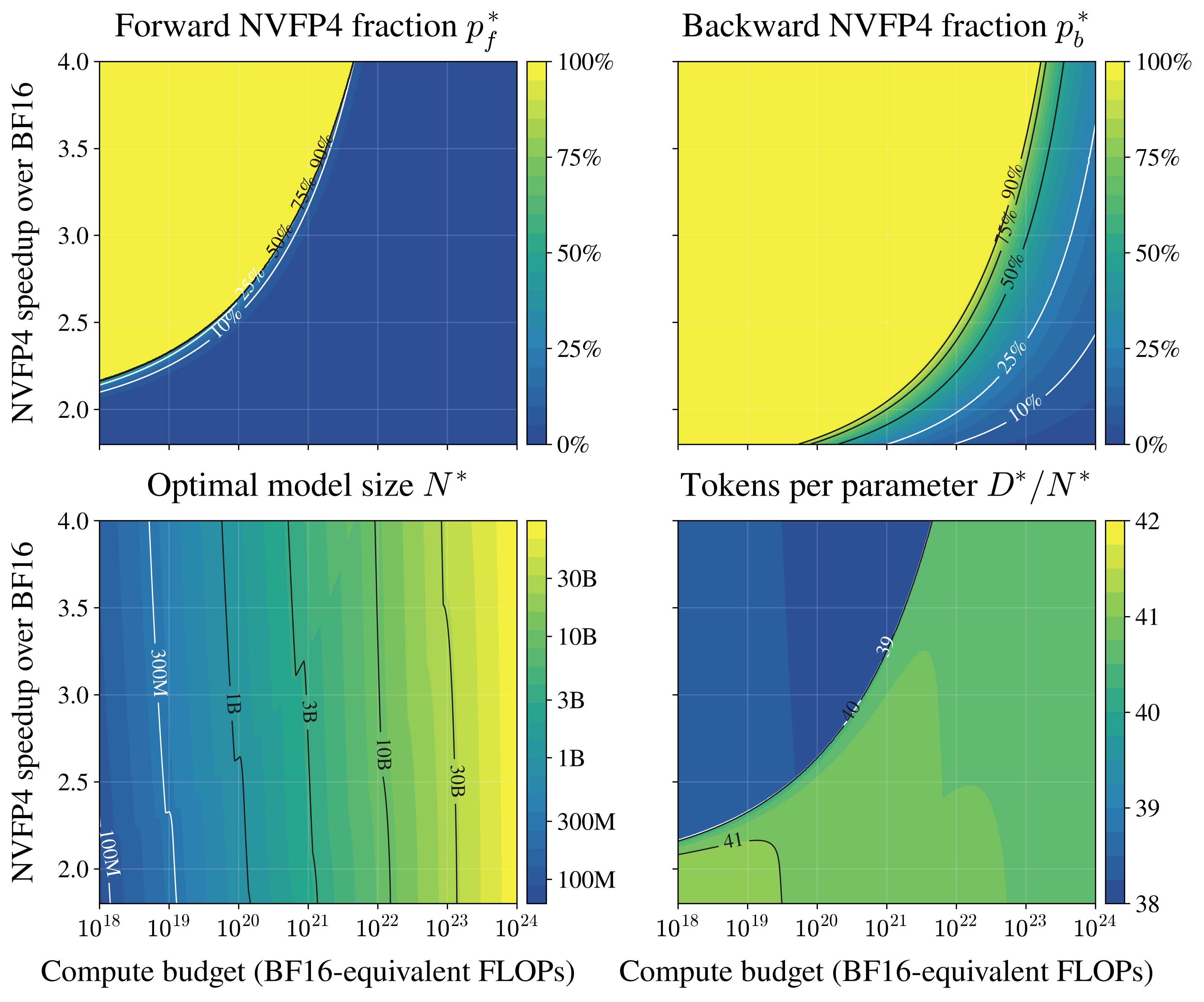}\\[-0.3em]
{\small (a) MXFP8 speedup $1.6\times$}\\[0.8em]
\includegraphics[width=0.7\textwidth]{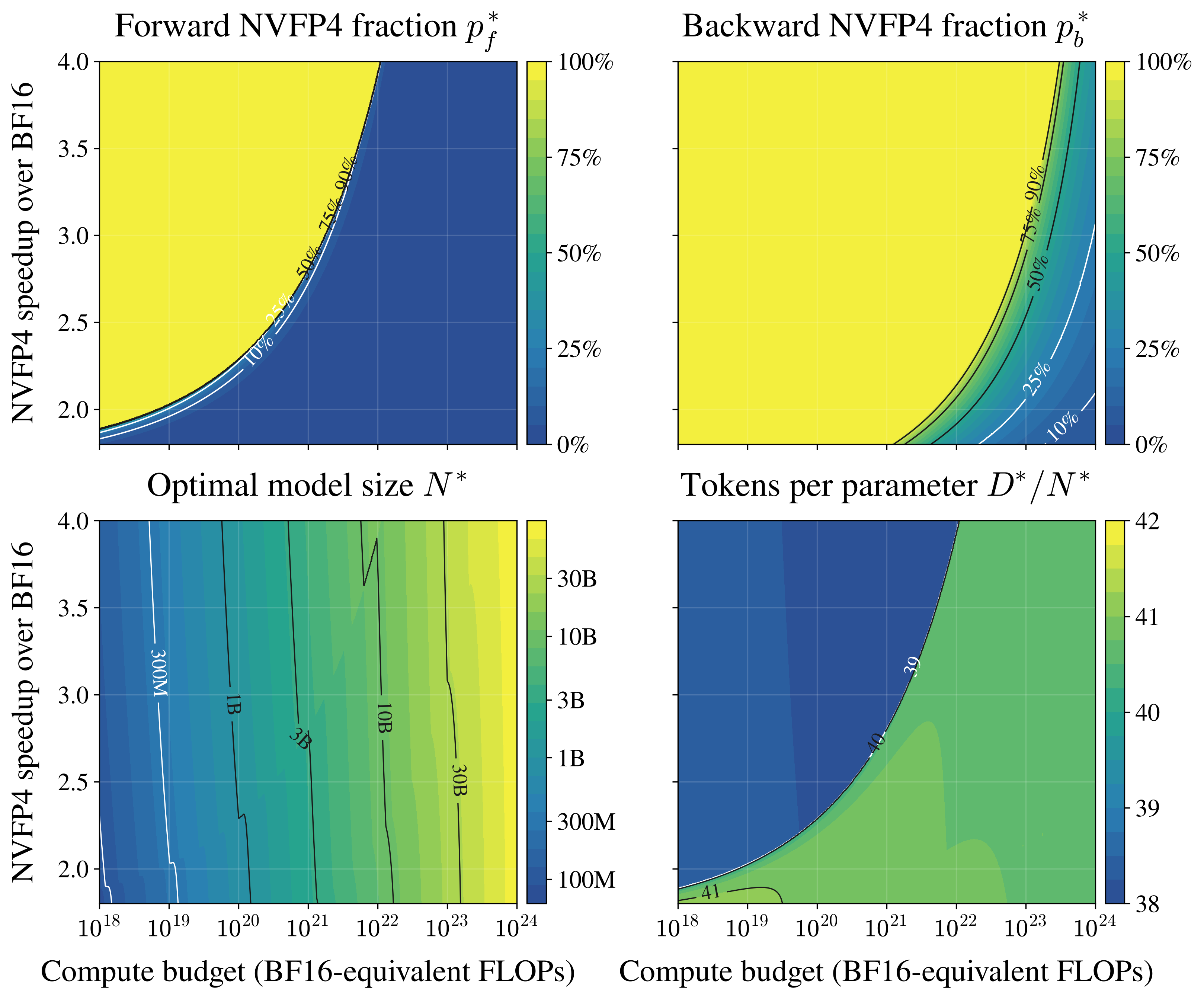}\\[-0.3em]
{\small (b) MXFP8 speedup $1.4\times$}
\caption{Compute-optimal precision allocation predicted by our scaling law, as in Figure~\ref{fig:compute-optimal}, with the MXFP8 speedup over BF16 set to (a) $1.6\times$ and (b) $1.4\times$ instead of $1.8\times$.}
\label{fig:compute-optimal-mxfp8}
\end{figure}

Figure~\ref{fig:compute-optimal} fixes the MXFP8 speedup over BF16 at $1.8\times$, following the GEMM kernel speedup reported by NVIDIA Transformer Engine \citep{transformerengine_gemm}, but end-to-end speedups can be lower in practice. Figure~\ref{fig:compute-optimal-mxfp8} shows the results with MXFP8 speedups of $1.6\times$ and $1.4\times$. A lower MXFP8 speedup makes higher precision more expensive, so NVFP4 remains optimal up to larger compute budgets. 
The qualitative conclusions of Section~\ref{sec:compute-optimal} hold in all cases, as the optimal precision still increases with the compute budget and the backward pass retains NVFP4 up to larger budgets than the forward pass, while the optimal model size and tokens per parameter remain nearly unchanged.

\section{Related work}
\label{sec:app-related}

\citet{kaplan2020scaling} propose a coupled power law of the form $L=(AN^{-\alpha}+BD^{-1})^{\gamma}$, where the outer exponent $\gamma$ couples the model and data terms and the inner data exponent is fixed to one. The Chinchilla scaling law \citep{hoffmann2022training} takes the additive form $L=AN^{-\alpha}+BD^{-\beta}+E$, allowing separate exponents for model size and data but no interaction between their contributions. More recent work combines independent inner exponents with an outer coupling exponent in the form $L=(AN^{-\alpha}+BD^{-\beta})^{\gamma}+E$, empirically improving loss prediction across data regimes \citep{busbridge2025distillation,videau2026skaling}.

Table~\ref{tab:scaling-law-formulations} summarizes existing laws that incorporate quantization or compression. For post-training quantization (PTQ), \citet{kumar2024scalingprecision} add a degradation term $C_T D^{\gamma_D}N^{-\gamma_N}e^{-P_{\mathrm{post}}/\gamma_{\mathrm{post}}}$ to the Chinchilla law, where $P_{\mathrm{post}}$ denotes the post-training precision. The degradation grows with the training token count and decreases with model size and precision. \citet{cao2024mixedquant} model mixed-precision PTQ by adding a degradation term $Ce^{\kappa p}N^{-\gamma_N}(G+d)^{\gamma_G}$, where $p$ is the fraction of parameters in low precision, chosen by random search, and $G$ is the block size. For quantized pretraining (QPT), \citet{kumar2024scalingprecision} replace $N$ with the effective parameter count $(1-e^{-P/\bar{\gamma}})^3N$, where $P$ denotes the training precision and $\bar{\gamma}$ is a fitted constant. \citet{frantar2025compression} generalize this rescaling to an effective parameter count $\rho_N(\psi_f)N$ that depends on the compression scheme. For quantization-aware training (QAT), \citet{chen2025qatscaling} add an error term $kD^{\gamma_D}N^{-\gamma_N}(\log_2 G)^{\gamma_G}$ that also depends on the quantization group size $G$. \citet{panferov2026unified} predict parameter efficiency from the representation error of the compression scheme, where $\mathrm{GMSE}(\psi_f)$ measures the error of fitting Gaussian data using the forward scheme $\psi_f$ and $\widetilde{\rho}$ maps this error to parameter efficiency. Finally, the Quartet law \citep{castro2025quartet} rescales both the parameter count and the training token count, with $\rho_N(\psi_f)$ depending on the forward scheme and $\rho_D(\psi_b)$ on the backward scheme.

\end{document}